\documentclass[11pt]{article}

\usepackage[preprint]{acl}

\usepackage{times}
\usepackage{latexsym}

\usepackage[T1]{fontenc}

\usepackage[utf8]{inputenc}

\usepackage{microtype}

\usepackage{inconsolata}

\usepackage{graphicx}

\usepackage{algorithm}
\usepackage{algpseudocode}
\usepackage{amsmath}
\usepackage{amsfonts}
\usepackage{booktabs}
\usepackage{enumitem} 
\usepackage{multirow}
\usepackage{graphicx} 
\usepackage[table]{xcolor}
\usepackage[most]{tcolorbox} 
\tcbset{
  promptstyle/.style={
    colback=gray!10,        
    colframe=black!40,      
    colbacktitle=gray!20,   
    coltitle=black,         
    boxrule=0.4pt,          
    arc=2pt,                
    outer arc=2pt,
    left=5pt, right=5pt, top=5pt, bottom=5pt,
    fonttitle=\bfseries,    
    title={#1}              
  }
}

\title{Memory-R1: Enhancing Large Language Model Agents to Manage and Utilize Memories via Reinforcement Learning}

\author{
 \textbf{Sikuan Yan\textsuperscript{*1,2}},
 \textbf{Xiufeng Yang\textsuperscript{*3}},
 \textbf{Zuchao Huang\textsuperscript{1}},
 \textbf{Ercong Nie\textsuperscript{1,2}},
\\
 \textbf{Zifeng Ding\textsuperscript{2,4}},
 \textbf{Zonggen Li\textsuperscript{5}},
 \textbf{Xiaowen Ma\textsuperscript{1}},
 \textbf{Jinhe Bi\textsuperscript{1}},
 \textbf{Kristian Kersting\textsuperscript{6}},
\\
 \textbf{Jeff Z. Pan\textsuperscript{7}},
 \textbf{Hinrich Schuetze\textsuperscript{1,2}},
 \textbf{Volker Tresp\textsuperscript{1,2}},
 \textbf{Yunpu Ma\textsuperscript{\dag}\textsuperscript{1,2}}
\\
 \textsuperscript{1}Ludwig Maximilian University of Munich,
 \textsuperscript{2}Munich Center for Machine Learning,
\\
 \textsuperscript{3}Technical University of Munich,
 \textsuperscript{4}University of Cambridge,
 \textsuperscript{5}University of Hong Kong,
\\
 \textsuperscript{6}Technical University of Darmstadt,
 \textsuperscript{7}University of Edinburgh
\\
 \small{
    \href{mailto:email@domain}{s.yan@campus.lmu.de}, \href{mailto:email@domain}{cognitive.yunpu@gmail.com}
 }
}

\begin{document}
\maketitle
\begin{abstract}

Large Language Models (LLMs) have demonstrated impressive capabilities across a wide range of NLP tasks, but they remain fundamentally \textit{stateless}, constrained by limited context windows that hinder long-horizon reasoning. Recent efforts to address this limitation often augment LLMs with an external memory bank, yet most existing pipelines are static and heuristic-driven, lacking a learned mechanism for deciding what to store, update, or retrieve. We present Memory-R1, a reinforcement learning (RL) framework that equips LLMs with the ability to actively manage and utilize external memory through two specialized agents: a \textit{Memory Manager} that learns structured operations, including ADD, UPDATE, DELETE, and NOOP; and an \textit{Answer Agent} that pre-selects and reasons over relevant entries. Both agents are fine-tuned with outcome-driven RL (PPO and GRPO), enabling adaptive memory management with minimal supervision. With only 152 training QA pairs, Memory-R1 outperforms strong baselines and generalizes across diverse question types, three benchmarks (LoCoMo, MSC, LongMemEval), and multiple model scales (3B–14B).
\end{abstract}

\begin{figure*}[t]
\centering
\includegraphics[width=1\textwidth]{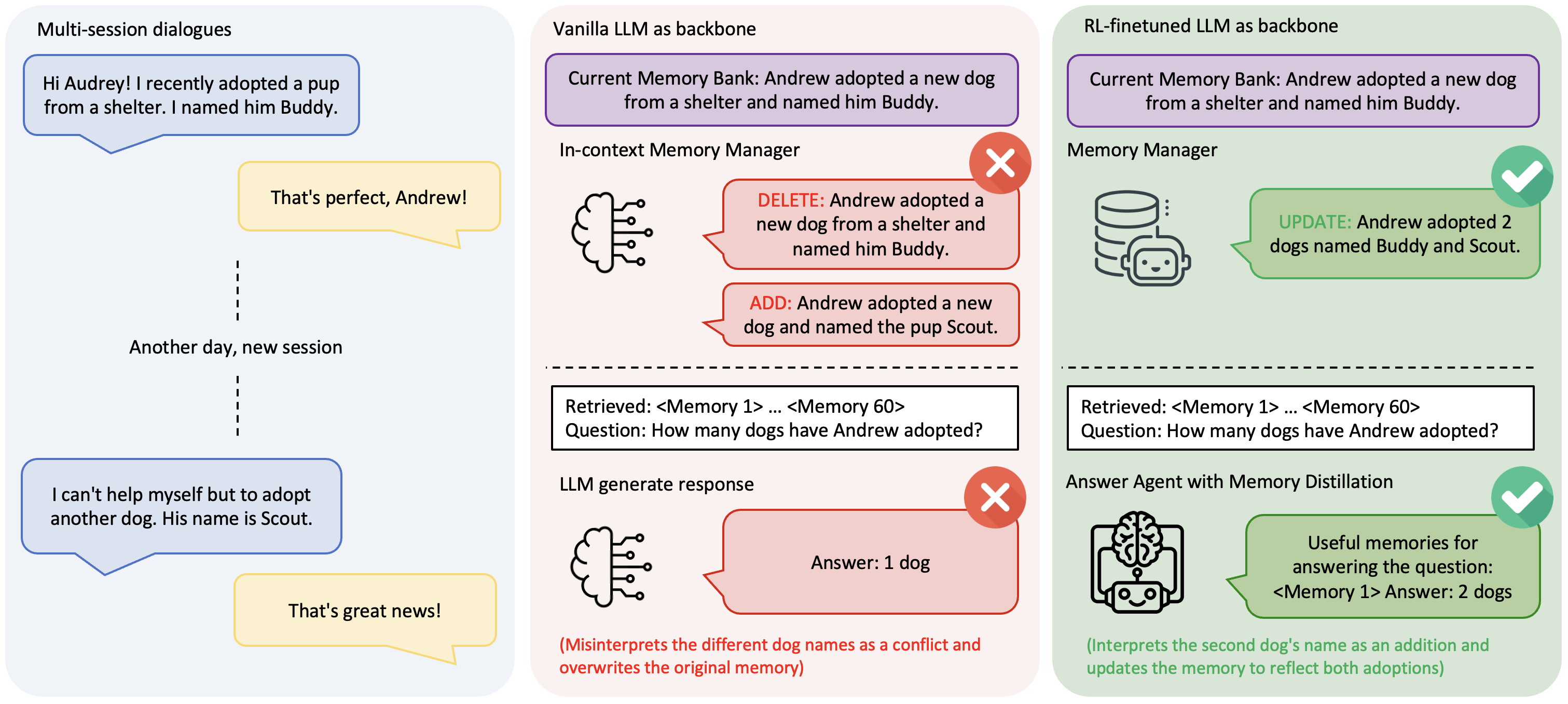} 

\caption{
Comparison of Memory-R1 and a vanilla LLM memory system. (Left) In a multi-session dialogue, the user mentions adopting two dogs across sessions. (Middle) The vanilla Memory Manager misinterprets this as a contradiction and issues DELETE+ADD, fragmenting memory. (Right) The RL-trained Memory Manager issues a single UPDATE to consolidate the fact, while the Answer Agent distills 60 retrieved memories down to the relevant one (“Andrew adopted 2 dogs named Buddy and Scout”) and correctly answers “2 dogs.”
}
\label{fig1-usecase}
\end{figure*}

\section{Introduction}
Large Language Models (LLMs) have shown remarkable ability in understanding and generating natural language, making them central to recent advances in AI~\citep{openai2024gpt4technicalreport, qwen2025qwen25technicalreport}. Yet, they remain fundamentally \textit{stateless}~\citep{yu2025stateful, fan2025if, goodyear2025effect}: their memory is bounded by a finite context window and any information that falls outside this window is forgotten, preventing them from maintaining knowledge across long conversations or evolving tasks~\citep{wang2024adaptingllmsefficientcontext, fei2023extendingcontextwindowlarge}.

One early effort is the Tensor Brain framework, which uses a bilayer tensor network with index and representation layers to model episodic, semantic, and working memory~\citep{tresp2023tensor}. Recent studies augment LLMs with explicit external memory modules~\citep{zhang2024survey}, most of which adopt the retrieval-augmented generation (RAG) paradigm~\citep{pan2025memory, salama2025meminsightautonomousmemoryaugmentation}, appending retrieved memory entries to the model’s input prompt. While this extends access to past information, it also creates a fundamental retrieval challenge: heuristics may return too few entries, omitting crucial context, or too many, flooding the model with irrelevant information and degrading performance~\citep{liu2023lostmiddlelanguagemodels}. In this paradigm, retrieved memories are passed to the LLM without meaningful filtering or prioritization, forcing the model to reason over both relevant and irrelevant content, which makes it prone to distraction by noise. Humans, by contrast, retrieve broadly but then filter, integrating only the most useful pieces to maintain coherent, evolving knowledge.

Equally important is the challenge of \textit{memory management}: deciding what to remember, update, or discard. Some systems~\citep{packer2023memgpt, modarressi2024memllm, xiong2025memory} adopt CRUD-style operations, namely create, read, update, and delete, which are adapted from databases~\citep{martin1983database}. A more recent work~\citep{aios2024sdk} augments this paradigm with a search operator, while Mem0~\citep{chhikara2025mem0} investigates the operator set \{\texttt{ADD}, \texttt{UPDATE}, \texttt{DELETE}, \texttt{NOOP}\}. We adopt this setting, as it provides a minimal yet expressive framework for modeling memory dynamics. Existing approaches mainly rely on vanilla LLMs to choose operations from in-context instructions without any learning signal tied to correctness~\citep{packer2023memgpt, chhikara2025mem0}. Even simple cases can fail. Figure \ref{fig1-usecase}, a simplified example drawn from a LoCoMo conversation~\citep{maharana2024evaluating}, shows how a user says \textit{“I adopted a dog named Buddy”} and later adds \textit{“I adopted another dog named Scout”}. A vanilla system misinterprets this as a contradiction, issuing \texttt{DELETE}+\texttt{ADD} and overwriting the original memory. A trained agent instead consolidates with an \texttt{UPDATE}: \textit{“Andrew adopted two dogs, Buddy and Scout.”} Appendix~\ref{app:case_study_memory_manager} provides a real dialogue trace illustrating this case in practice.

These challenges of retrieving and managing memory remain largely unsolved. Supervised fine-tuning provides limited help because it is impractical to label every memory operation or retrieval decision. Reinforcement learning (RL), in contrast, has recently shown strong potential for aligning LLM behavior with high-level objectives, including tool use~\citep{qian2025toolrlrewardtoollearning, wang2025actingreasoningmoreteaching}, web navigation~\citep{wei2025webagentr1trainingwebagents}, and search optimization~\citep{jin2025search, song2025r1}. Building on this success, we argue that RL is the missing ingredient for adaptive memory in LLM agents. By optimizing outcome-based rewards, models can learn when to add, update, delete, or retain information and how to use retrieved memories for reasoning.

In this paper, we present Memory-R1, an RL fine-tuned, memory-augmented LLM framework with two specialized agents: (1) a \textit{Memory Manager} that performs structured memory operations to maintain and evolve the memory bank, and (2) an \textit{Answer Agent} that applies a \textit{Memory Distillation} policy to filter memories retrieved via Retrieval‑Augmented Generation (RAG) and reason over the selected entries to produce answers. Both agents are fine-tuned using PPO~\citep{schulman2017proximal} or GRPO~\citep{shao2024deepseekmath}, achieving strong performance with as few as 152 question–answer pairs. On the LoCoMo benchmark~\citep{maharana2024evaluating}, Memory-R1 delivers substantial gains over the most competitive baseline, Mem0~\citep{chhikara2025mem0}. Using the LLaMA-3.1-8B-Instruct backbone, Memory-R1-GRPO achieves relative improvements of 28\% in F1, 34\% in BLEU-1, and 30\% in LLM-as-a-Judge. These improvements set a new state of the art on LoCoMo and underscore Memory‑R1’s ability to achieve large performance gains with minimal supervision, highlighting its efficiency. 

Our contributions are summarized as follows: (1) We introduce Memory-R1, the first RL framework for memory-augmented LLMs, consisting of a \textit{Memory Manager} to perform structured memory operations and an \textit{Answer Agent} to filter and reason over memories retrieved via RAG. (2) We develop a data-efficient fine-tuning strategy using PPO and GRPO that enables Memory-R1 to achieve strong performance with as few as 152 question–answer pairs, demonstrating that large memory improvements can be achieved with minimal supervision. (3) We provide in-depth analysis of RL choices, model size, and memory design, offering actionable insights for building the next generation of memory-aware, reasoning-capable LLM agents.

\section{Related Work}
\subsection{Memory Augmented LLM-based Agents}
LLMs have emerged as powerful general-purpose reasoners, capable of engaging in multi-turn dialogues, decomposing tasks into actionable steps, and leveraging prior context to guide decision making~\citep{brown2020languagemodelsfewshotlearners,chowdhery2022palmscalinglanguagemodeling,openai2024gpt4technicalreport}. However, their reliance on fixed-length context windows limits their ability to retain information over extended interactions. To overcome this, recent work augments LLM agents with external memory modules, enabling long-horizon reasoning and persistent knowledge accumulation through selective storage, retrieval, and updating of information. Several approaches illustrate this trend. LoCoMo~\citep{maharana2024evaluating} introduces a benchmark to evaluate agents’ ability to retrieve and reason over temporally distant conversational history. ReadAgent~\citep{lee2024human} proposes a human-inspired reading agent that uses gist-based memory for reasoning over very long contexts.
MemoryBank~\citep{zhong2024memorybank} proposes a compositional memory controller for lifelong agent memory. MemGPT~\citep{packer2023memgpt} introduces working and long-term buffers with scheduling policies. For a broader perspective, we refer readers to the recent survey on memory systems in AI agents~\citep{du2025rethinkingmemoryaitaxonomy}. While most existing approaches rely on static memory designs, our work instead develops a learnable memory system trained with reinforcement learning.

\subsection{LLM and Reinforcement Learning}
The intersection of LLM and RL has received increasing attention as researchers seek to move beyond static supervised fine-tuning and enable models to learn from dynamic, interactive feedback. Reinforcement Learning from Human Feedback (RLHF)~\citep{ouyang2022traininglanguagemodelsfollow} is a foundational method used to align LLM outputs with human preferences. Recent works extend RL to more structured decision-making tasks for LLMs. For instance, Toolformer~\citep{schick2023toolformerlanguagemodelsteach} and ReAct-style agents~\citep{yao2023reactsynergizingreasoningacting} frame tool use as an RL problem, where the LLM learns when to query external tools or APIs. Search-R1~\citep{jin2025search} trains LLMs to issue web search queries using RL to maximize final answer correctness. Similarly, the Trial and Error approach~\citep{song2024trialerrorexplorationbasedtrajectory} optimizes agents to select better reasoning paths. These approaches demonstrate that RL can improve complex behavior sequences in LLMs. However, memory management and utilization in LLMs remain underexplored in the RL setting. Existing memory-augmented LLM systems~\citep{chhikara2025mem0, packer2023memgpt} typically rely on heuristics to control memory operations, lacking adaptability and long-term optimization. Our work, Memory-R1, is among the first to frame memory operation selection, and the utilization of relevant memories as an RL problem.

\begin{figure*}[t]
\centering
\includegraphics[width=1\textwidth]{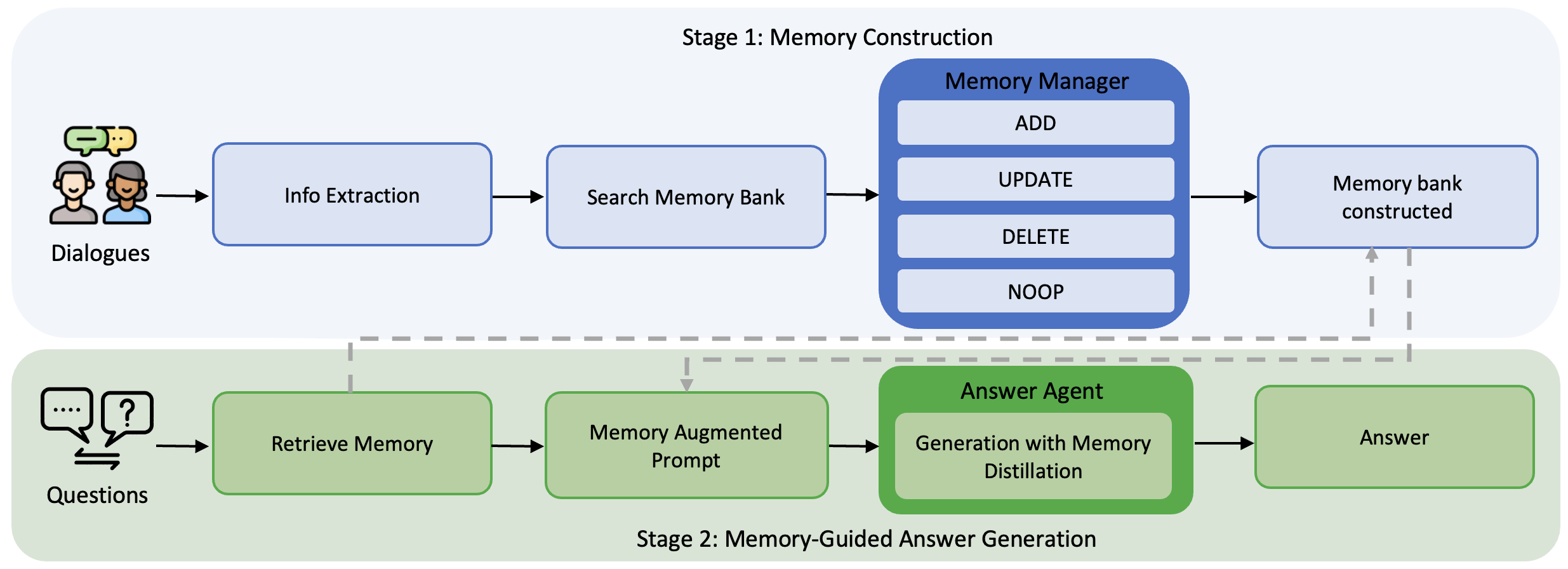} 
\caption{
Overview of the Memory-R1 framework. Stage 1 (blue) constructs and updates the memory bank via the RL‑fine‑tuned Memory Manager, which chooses operations \{ADD, UPDATE, DELETE, NOOP\} for each new dialogue turn. Stage 2 (green) answers user questions via the Answer Agent, which applies a Memory Distillation policy to reason over retrieved memories.
}
\label{figure2_pipeline}
\end{figure*}

\section{Method}

We present Memory-R1, a reinforcement learning framework for multi-session dialogue tasks, where each dialogue contains multiple \textit{sessions} (separate interactions occurring at different times) and each session consists of several \textit{turns} (a back-and-forth exchange between two users). Answering a question always requires synthesizing information spread across sessions, posing a strong challenge for long-horizon memory management and reasoning. Figure~\ref{figure2_pipeline} illustrates the overall pipeline. At each dialogue turn, the LLM extracts and summarizes information worth remembering, then retrieves related entries from the memory bank as part of the Retrieval-Augmented Generation (RAG) framework. The \textit{Memory Manager} decides whether to \texttt{ADD}, \texttt{UPDATE}, \texttt{DELETE}, or \texttt{NOOP}, thereby maintaining and evolving the memory state. For question answering, the \textit{Answer Agent} applies a memory distillation policy over retrieved memories to filter noise and reason over the most relevant content. Both agents are fine-tuned with PPO or GRPO, enabling outcome-driven learning of memory operations and selective utilization. Further implementation details, such as model hyperparameters, optimization schedule, and training setup, are provided in Appendix~\ref{appendix:implementation-details}.

\subsection{RL Fine-tuning for Memory Manager}  
\label{RL_for_memory_manager}
\paragraph{Task Formulation}
The \textit{Memory Manager} maintains the memory bank by selecting one of {\texttt{ADD}, \texttt{UPDATE}, \texttt{DELETE}, \texttt{NOOP}} for each new piece of information extracted from a dialogue, outputting both the operation and updated content $m'$. Training uses (i) a partially constructed memory bank and (ii) a new dialogue turn with information relevant to downstream QA. The goal is to learn which operation produces a memory state that enables the Answer Agent to answer correctly. Formally, the Memory Manager is modeled as a policy $\pi_\theta$ that takes extracted information $x$ and retrieved memories $\mathcal{M}_{\text{old}}$ as input, and outputs an operation $o$ with content $m'$:
\begin{equation}
(o, m') \sim \pi\theta(\cdot \mid x, \mathcal{M}_{\text{old}}),
\end{equation}
where $x$ is the new information and $\mathcal{M}_{\text{old}}$ the current memory bank. The data construction details are provided in Appendix~\ref{sec:data_construction}.

\paragraph{PPO for Memory Manager}  
We fine-tune the Memory Manager with \textit{Proximal Policy Optimization} (PPO;~\citealp{schulman2017proximal}).  
Given candidate memory $x$ and memory bank $\mathcal{M}_{\text{old}}$, the manager samples an operation $o$ and updated content $m'$ from policy $\pi_\theta$, applies it to the memory bank, and forwards the result to the frozen Answer Agent.  
Answer correctness provides a scalar reward $r$, from which we estimate an advantage $A$.  
The clipped PPO objective is:

\begin{equation}
\label{eq:ppo_equation}
\mathcal{J}(\theta) =
\mathbb{E} \left[
\min\big( \rho_\theta A,\;
\text{clip}(\rho_\theta, 1-\epsilon, 1+\epsilon) A \big)
\right],
\end{equation}

\noindent where 
$\rho_\theta = 
\frac{\pi_\theta(o, m' \mid x, \mathcal{M}_{\text{old}})}
{\pi_{\text{old}}(o, m' \mid x, \mathcal{M}_{\text{old}})}
$
is the importance ratio, $A$ is the advantage estimated from the answer-based reward $r$, and $\epsilon$ is the clipping threshold for stable updates.  

\paragraph{GRPO for Memory Manager}  
We also train the Memory Manager with \textit{Group Relative Policy Optimization} (GRPO;~\citealp{shao2024deepseekmath}), which samples a group of $G$ candidate actions per state and computes their relative advantages.  
This formulation avoids an explicit value function while maintaining PPO-style stability.  
For a state $s=(x,\mathcal{M}_{\text{old}})$, the GRPO objective is:

\begin{equation}
\label{eq:grpo_equation}
\mathcal{J}(\theta) = 
\mathbb{E} \Bigg[ 
\frac{1}{G} \sum_{i=1}^G 
\rho_\theta^{(i)} A_i
- \beta \, \mathbb{D}_{\text{KL}} \!\left[ \pi_\theta \,\|\, \pi_{\text{ref}} \right]
\Bigg],
\end{equation}

\noindent where each candidate $i$ yields reward $r_i$,  
$
A_i = \frac{r_i - \text{mean}(\mathbf{r})}{\text{std}(\mathbf{r})}, 
\quad \mathbf{r} = \{r_1,\ldots,r_G\},
$ 
is its standardized group-relative advantage, and $\rho_\theta^{(i)}$ is the per-action importance ratio.  
The KL term regularizes updates to prevent policy drift away from the reference $\pi_{\text{ref}}$.

\paragraph{Reward Design for Memory Manager}  
We use an outcome-driven reward: the Memory Manager’s operations are judged by their effect on downstream QA. After applying operation $o$ with proposed content $m'$, the updated memory bank is passed to the frozen Answer Agent, and the reward is based on answer correctness:  

\begin{equation}
R_{answer} = \mathrm{EM}(y_{\text{pred}}, y_{\text{gold}})
\end{equation}

\noindent where $y_{\text{pred}}$ is the predicted answer and $y_{\text{gold}}$ the ground truth. This exact-match signal requires no manual labels, remains scalable, and is sufficient to teach effective memory operations.

\subsection{RL Fine-Tuning for Answer Agent}
\label{RL_for_answer agent}
\paragraph{Task Formulation}  
The Answer Agent leverages the memory bank maintained by the Memory Manager to answer questions in multi-session dialogues. Following~\citep{chhikara2025mem0}, 60 candidate memories are retrieved for each question via similarity-based RAG, and the agent performs \textit{memory distillation} to select the most relevant entries before generating an answer.  

We model the agent as a policy $\pi_\theta$ mapping the question $q$ and retrieved set $\mathcal{M}_{\text{ret}}$ to an answer $y$:  
\begin{equation}
y \sim \pi_\theta(\cdot \mid q, \mathcal{M}_{\text{ret}}).
\end{equation}

\paragraph{PPO for Answer Agent}  
We fine‑tune the Answer Agent using the same PPO algorithm as in Section~\ref{RL_for_memory_manager}. The agent takes the question $q$ and retrieved memories $\mathcal{M}_{\text{ret}}$ and generates an answer $y$. The objective mirrors Equation~\eqref{eq:ppo_equation}, applied to the generated sequence. The importance ratio is:  
\begin{equation}
\rho_{\theta}(q, \mathcal{M}_{\text{ret}}) =
\frac{\pi_{\theta}(y \mid q, \mathcal{M}_{\text{ret}})}
{\pi_{\text{old}}(y \mid q, \mathcal{M}_{\text{ret}})} ,
\end{equation}
\noindent where $y$ is the generated answer. Advantages derive from final answer quality (e.g., exact match), and clipping ensures stable updates.  

\paragraph{GRPO for Answer Agent}  
We also fine‑tune the Answer Agent with GRPO, following the formulation in Section~\ref{RL_for_memory_manager}. For each $(q, \mathcal{M}_{\text{ret}})$, the policy samples $G$ candidate answers $\{y_i\}_{i=1}^G$. Their exact-match rewards against $y_{\text{gt}}$ are normalized into group-relative advantages. GRPO uses the same importance ratio as PPO, computed per candidate, and stabilizes training without a value function by comparing candidates within each group.

\paragraph{Reward Design for Answer Agent}
We use the Exact Match (EM) score between the generated answer $y_{\text{pred}}$ and ground truth $y_{\text{gold}}$ as the reward. This design directly ties the reward to the correctness of the final answer, encouraging the agent to select and reason over memories in a way that yields accurate outputs rather than optimizing for intermediate steps.




\begin{table*}[t]
\centering
\scalebox{0.65}{
\begin{tabular}{l|l|ccc|ccc|ccc|ccc|ccc}
\toprule
\textbf{Model} & \textbf{Method} & 
\multicolumn{3}{c|}{\textbf{Single Hop}} & 
\multicolumn{3}{c|}{\textbf{Multi-Hop}} & 
\multicolumn{3}{c|}{\textbf{Open Domain}} & 
\multicolumn{3}{c|}{\textbf{Temporal}} & 
\multicolumn{3}{c}{\textbf{Overall}} \\
& & F1$\uparrow$ & B1$\uparrow$ & J$\uparrow$ 
& F1$\uparrow$ & B1$\uparrow$ & J$\uparrow$ 
& F1$\uparrow$ & B1$\uparrow$ & J$\uparrow$ 
& F1$\uparrow$ & B1$\uparrow$ & J$\uparrow$ 
& F1$\uparrow$ & B1$\uparrow$ & J$\uparrow$ \\
\midrule

\multirow{6}{*}{\rotatebox{90}{%
    \begin{tabular}{c}
        \textbf{LLaMA-3.1-8B} \\
        \textbf{Instruct}
    \end{tabular}}}
& LoCoMo (RAG)   & 12.25 & 9.77 & 13.81 & 13.69 & 10.96 & 20.48 & 11.59 & 8.30 & 15.96 & 9.38 & 8.15 & 4.65 & 11.41 & 8.71 & 13.62 \\
& A-Mem          & 21.62 & 16.93 & 44.76 & 13.82 & 11.45 & 34.93 & 34.67 & 29.13 & 49.38 & 25.77 & 22.14 & 36.43 & 29.20 & 24.40 & 44.76 \\
& Mem0           & 27.29 & 18.63 & 43.93 & 18.59 & 13.86 & 37.35 & 34.03 & 24.77 & 52.27 & 26.90 & 21.06 & 31.40 & 30.41 & 22.22 & 45.68 \\
& MemoryOS            & 31.89 & 23.05 & 52.72 & 13.80 & 12.78 & 31.33 & 40.74 & 33.67 & 57.36 & 28.74 & 21.44 & 23.64 & 35.04 & 27.99 & 48.20 \\
& Memory-SFT & 34.64 & 23.73 & 56.90 & 20.80 & 16.26 & 37.35 & 46.47 & 37.35 & 63.27 & 47.18 & 34.58 & \textbf{54.65} & 42.81 & 32.98 & 58.76 \\
& \cellcolor{gray!15}Memory-R1-PPO  & \cellcolor{gray!15}32.52 & \cellcolor{gray!15}24.47 & \cellcolor{gray!15}53.56 & \cellcolor{gray!15}26.86 & \cellcolor{gray!15}23.47 & \cellcolor{gray!15}42.17 & \cellcolor{gray!15}45.30 & \cellcolor{gray!15}39.18 & \cellcolor{gray!15}64.10 & \cellcolor{gray!15}41.57 & \cellcolor{gray!15}26.11 & \cellcolor{gray!15}47.67 & \cellcolor{gray!15}41.05 & \cellcolor{gray!15}32.91 & \cellcolor{gray!15}57.54 \\
& \cellcolor{gray!15}Memory-R1-GRPO & \cellcolor{gray!15}\textbf{35.73} & \cellcolor{gray!15}\textbf{27.70} & \cellcolor{gray!15}\textbf{59.83} & \cellcolor{gray!15}\textbf{35.65} & \cellcolor{gray!15}\textbf{30.77} & \cellcolor{gray!15}\textbf{53.01} & \cellcolor{gray!15}\textbf{47.42} & \cellcolor{gray!15}\textbf{41.24} & \cellcolor{gray!15}\textbf{68.78} & \cellcolor{gray!15}\textbf{49.86} & \cellcolor{gray!15}\textbf{38.27} & \cellcolor{gray!15}51.55 & \cellcolor{gray!15}\textbf{45.02} & \cellcolor{gray!15}\textbf{37.51} & \cellcolor{gray!15}\textbf{62.74} \\

\cmidrule(lr){1-17}

\multirow{6}{*}{\rotatebox{90}{%
    \begin{tabular}{c}
        \textbf{Qwen-2.5-7B} \\
        \textbf{Instruct}
    \end{tabular}}}
& LoCoMo (RAG)   & 9.57 & 7.00 & 15.06 & 11.84 & 10.02 & 19.28 & 8.67 & 6.52 & 12.79 & 8.35 & 8.74 & 5.43 & 8.97 & 7.27 & 12.17 \\

& A-Mem          & 18.96 & 12.86 & 40.78 & 14.73 & 12.66 & 31.32 & 30.58 & 26.14 & 46.90 & 23.67 & 20.67 & 28.68 & 26.08 & 21.78 & 40.78 \\
& Mem0           & 24.96 & 18.05 & 61.92 & 20.31 & 15.82 & 48.19 & 32.74 & 25.27 & 65.20 & 33.16 & 26.28 & 38.76 & 30.61 & 23.55 & 53.30 \\
& MemoryOS            & 29.55 & 22.59 & 48.12 & 21.03 & 18.41 & 38.55 & 40.85 & 36.26 & 63.14 & 26.26 & 19.70 & 24.81 & 34.64 & 29.36 & 51.26 \\
& Memory-SFT & 27.81 & 20.25 & 57.74 & 24.62 & 22.28 & 46.99 & 43.33 & 34.06 & 66.85 & 44.41 & 34.32 & \textbf{52.71} & 39.51 & 30.84 & 61.13 \\
& \cellcolor{gray!15}Memory-R1-PPO  & \cellcolor{gray!15}\textbf{34.22} & \cellcolor{gray!15}23.61 & \cellcolor{gray!15}57.74 & \cellcolor{gray!15}\textbf{32.87} & \cellcolor{gray!15}\textbf{29.48} & \cellcolor{gray!15}\textbf{53.01} & \cellcolor{gray!15}44.78 & \cellcolor{gray!15}38.72 & \cellcolor{gray!15}66.99 & \cellcolor{gray!15}42.88 & \cellcolor{gray!15}30.30 & \cellcolor{gray!15}42.25 & \cellcolor{gray!15}41.72 & \cellcolor{gray!15}33.70 & \cellcolor{gray!15}59.53 \\
& \cellcolor{gray!15}Memory-R1-GRPO & \cellcolor{gray!15}33.64 & \cellcolor{gray!15}\textbf{26.06} & \cellcolor{gray!15}\textbf{62.34} & \cellcolor{gray!15}23.55 & \cellcolor{gray!15}20.71 & \cellcolor{gray!15}40.96 & \cellcolor{gray!15}\textbf{46.86} & \cellcolor{gray!15}\textbf{40.92} & \cellcolor{gray!15}\textbf{67.81} & \cellcolor{gray!15}\textbf{47.75} & \cellcolor{gray!15}\textbf{38.49} & \cellcolor{gray!15}49.61 & \cellcolor{gray!15}\textbf{43.14} & \cellcolor{gray!15}\textbf{36.44} & \cellcolor{gray!15}\textbf{61.51} \\
\bottomrule
\end{tabular}}
\caption{Evaluation results of Memory-R1 and baselines across LLaMA‑3.1‑8B‑Instruct and Qwen‑2.5‑7B‑Instruct on the LoCoMo benchmark dataset. Models are evaluated on F1, BLEU‑1 (B1), and LLM‑as‑a‑Judge (J) across \textit{Single Hop}, \textit{Multi‑Hop}, \textit{Open Domain}, and \textit{Temporal} questions. Higher values indicate better performance. The best results are marked in bold.}
\label{tab:main-result-unified}
\end{table*}

\section{Experiments}
\label{section:experiments}

\subsection{Experimental Setup}

\paragraph{Dataset and Model}
We evaluate Memory-R1 on three benchmarks: LoCoMo~\citep{maharana2024evaluating}, MSC~\citep{packer2023memgpt}, and LongMemEval~\citep{wu2024longmemeval}. LoCoMo contains long multi-session dialogues (about 600 turns, 26k tokens) with QA pairs covering single-hop, multi-hop, open-domain, and temporal reasoning. Following prior work~\citep{chhikara2025mem0}, we exclude the adversarial subset and use a 1:1:8 train/validation/test split (152/81/1307 questions). Models are trained only on LoCoMo and evaluated zero-shot on MSC and LongMemEval. We use LLaMA-3.1-8B-Instruct and Qwen-2.5 Instruct backbones (3B, 7B, 14B). Dataset construction details are provided in Appendix~\ref{sec:dataset_details}.

\paragraph{Evaluation Metrics}
We evaluate performance using three metrics: token-level F1 (F1), BLEU-1 (B1), and LLM-as-a-Judge (J). F1 and B1 measure lexical overlap with ground-truth answers, while J uses a separate LLM to assess semantic correctness, relevance, completeness, and contextual appropriateness. Implementation details for LLM-as-a-Judge are provided in Appendix~\ref{appendix_prompt}.

\paragraph{Baselines}  
To evaluate the effectiveness of \textsc{Memory‑R1}, we compare it against several established baselines for multi‑session dialogue reasoning: 
(1) LoCoMo~\citep{maharana2024evaluating}, a RAG-style framework that converts entire dialogues into chunks and retrieves relevant segments for answering questions, serving as the benchmark baseline for long-range, multi-session conversation reasoning;
(2) A‑Mem~\citep{xu2025amemagenticmemoryllm}, a dynamic agentic memory system that creates, links, and updates structured memories to enhance reasoning across sessions;
(3) Mem0~\citep{chhikara2025mem0}, a modular memory system with explicit in context memory operations designed for scalable deployment;
(4) MemoryOS~\citep{kang2025memory}, a system-level framework that treats memory as an operating system abstraction for LLMs, providing unified mechanisms for memory read, write, and management across sessions to support long-horizon reasoning;
(5) Memory-SFT. To isolate the effect of RL, we implement a supervised fine-tuning variant of our framework.
Memory-SFT uses the same architecture and training data as Memory-R1 but replaces RL optimization with behavior cloning from GPT-5-generated trajectories.

For a fair comparison, we re‑implemented all baselines using both the LLaMA‑3.1‑8B‑Instruct and Qwen‑2.5‑7B‑Instruct models as backbones, with temperature set to 0 and a maximum token limit of 2048. This consistent setup ensures reproducibility and allows us to assess how each method performs across different model architectures.

\begin{figure*}[t]
\centering
\includegraphics[width=1.0\textwidth]{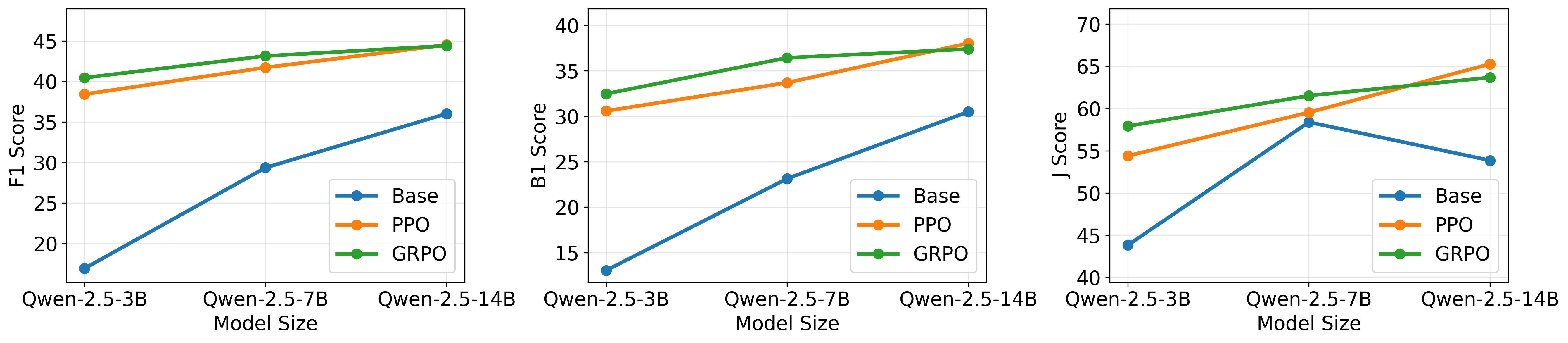} 
\caption{Scalability of Memory-R1 across model sizes (Qwen-2.5-3B, 7B, 14B-Instruct). Both PPO- and GRPO-tuned variants consistently outperform the base models across F1, BLEU-1 (B1), and LLM-as-a-Judge (J) metrics, showing strong scaling behavior.}
\label{fig:qwen_scaling}
\end{figure*}

\begin{figure}[t]
\centering
\includegraphics[width=1.0\columnwidth]{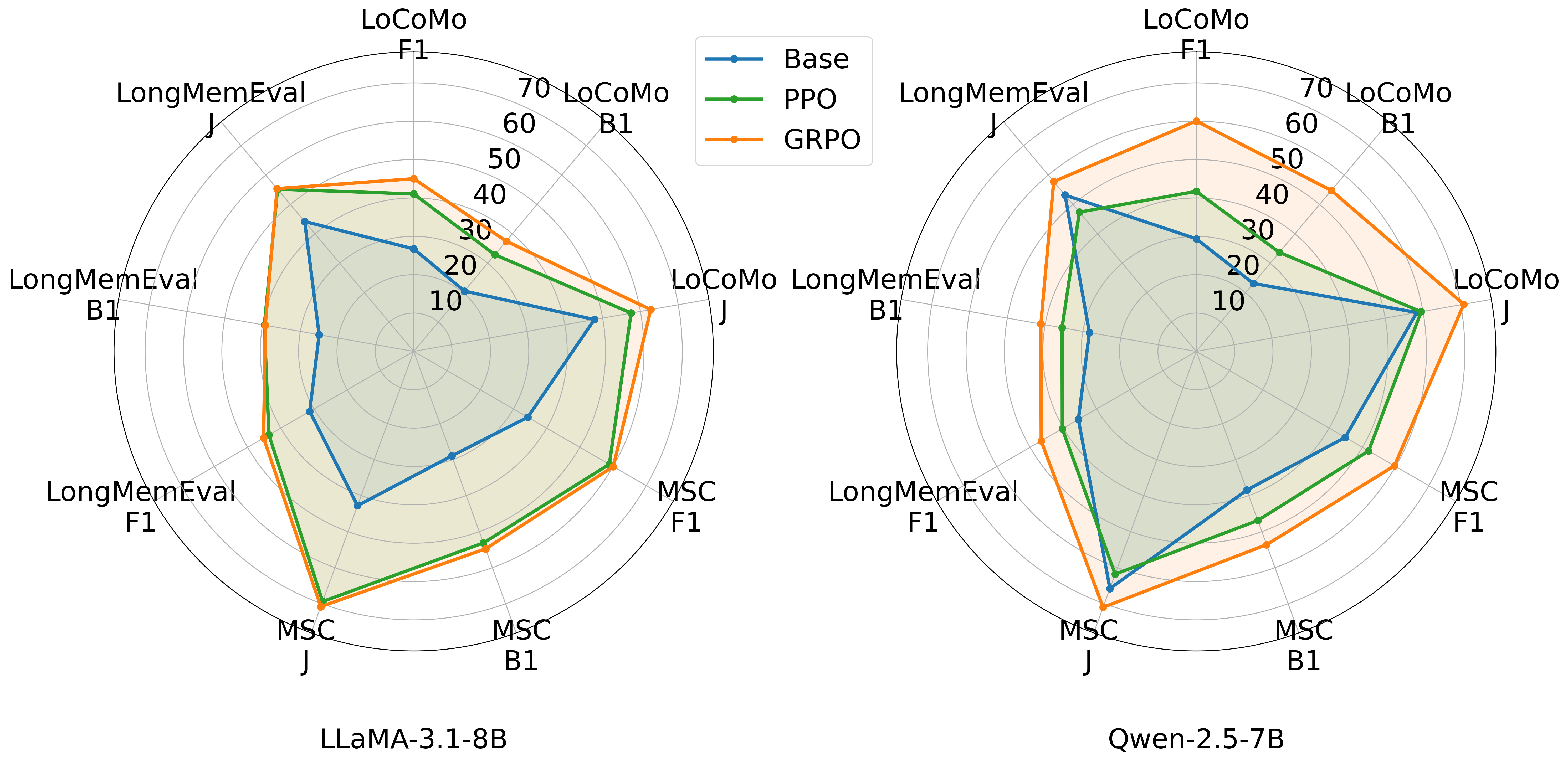} 
\caption{Generalization analysis of Memory-R1 across three benchmarks (LoCoMo, MSC, and LongMemEval), using LLaMA-3.1-8B-Instruct (left) and Qwen-2.5-7B-Instruct (right) as backbones.}
\label{fig:radar_three_datasets}
\end{figure}

\begin{figure*}[t]
\centering
\includegraphics[width=1\textwidth]{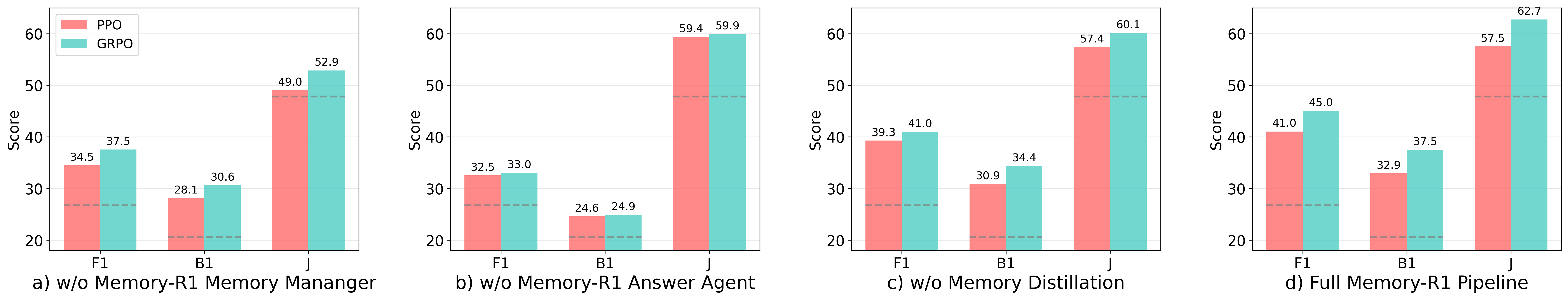} 
\caption{Ablation analysis of Memory-R1. Each subfigure shows the effect of removing one component: (a) Memory Manager, (b) Answer Agent, (c) Memory Distillation, and (d) the full pipeline. Performance drops in all ablations, demonstrating that each component contributes to the final results. Grey dashed lines indicate the baseline pipeline without RL fine-tuning.}
\label{fig:ablation_study}
\end{figure*}

\subsection{Main Results}
Table~\ref{tab:main-result-unified} reports the performance of Memory-R1 across LLaMA-3.1-8B-Instruct and Qwen-2.5-7B-Instruct models on the LoCoMo benchmark, covering diverse question types including single-hop, multi-hop, open-domain, and temporal reasoning. We evaluate two variants of Memory-R1, one fine-tuned with PPO and another with GRPO, and benchmark them against leading memory-augmented baselines, including LoCoMo (RAG), A-Mem, Mem0, MemoryOS and Memory-SFT.

Across both model families, Memory-R1 consistently achieves new state-of-the-art performance. On LLaMA-3.1-8B, Memory-R1-GRPO delivers the strongest overall performance, improving F1 by 28.5\%, B1 by 34.0\%, and J by 30.2\% relatively over the strongest baseline MemoryOS. Similarly, Memory-R1-PPO also yields substantial improvements, raising overall F1, B1, and J scores by 17.2\%, 17.6\%, and 19.4\%, respectively. When applied to Qwen-2.5-7B-Instruct, Memory-R1-GRPO again emerges as the top performer, surpassing MemoryOS by margins of 24.5\% (F1), 24.1\% (B1), and 20.0\% (J). PPO remains competitive, delivering strong gains over all non-RL baselines. Notably, while Memory-SFT benefits from guidance by a powerful teacher model (GPT-5), our reinforcement learning approach still outperforms it, highlighting the effectiveness of outcome-driven optimization over purely supervised imitation.

\subsection{Generalization and Scalability}
We further investigate the robustness of Memory-R1 across model scales and datasets. Figure~\ref{fig:qwen_scaling} shows results on the Qwen-2.5 family (3B, 7B, 14B). Memory-R1 consistently outperforms the base model at every scale, with PPO and GRPO delivering clear gains in F1, BLEU-1, and J scores. These improvements persist as models scale, demonstrating that reinforcement learning remains effective in teaching LLMs memory management regardless of backbone capacity.

To evaluate cross-task generalization, we apply the pipeline fine-tuned only on LoCoMo directly to two additional benchmarks: MSC and LongMemEval. As shown in Figure~\ref{fig:radar_three_datasets}, Memory-R1 with both PPO and GRPO continues to achieve consistent improvements across all three datasets and metrics, despite never being trained on MSC or LongMemEval. This zero-shot transfer highlights the robustness of Memory-R1 and shows its ability to generalize beyond its training distribution. The gains extend across single-hop, multi-hop, open-domain, and temporal questions, demonstrating Memory-R1 as a generalizable framework for adaptive, memory-augmented LLMs capable of long-horizon reasoning. Detailed results on LoCoMo, MSC, and LongMemEval, with type-level breakdowns, are provided in Appendix~\ref{appendix:extended_results}.

\subsection{Ablation Studies}
We conduct ablation studies to assess the contribution of each component in Memory-R1, isolating the effects of the Memory Manager, the Answer Agent, and the Memory Distillation mechanism. We also compare the training dynamics of PPO and GRPO.

\paragraph{Effect of Memory Manager}
We compare the full Memory-R1 pipeline with an ablated variant without RL fine-tuning of the Memory Manager, both using LLaMA-3.1-8B-Instruct. As shown in Figure~\ref{fig:ablation_study} (a,d), removing the RL-fine-tuned Memory Manager consistently degrades performance. Under PPO, F1, BLEU-1, and LLM-as-a-Judge drop from 41.0, 32.9, and 57.5 to 34.5, 28.1, and 49.0, respectively. Under GRPO, the corresponding scores decrease to 37.5, 30.6, and 52.9. These results confirm that outcome-driven RL enables more effective memory operations than scripted control.

\paragraph{Effect of Answer Agent}
Figure~\ref{fig:ablation_study} (b,d) shows that RL fine-tuning the Answer Agent substantially improves answer quality. Without the Memory-R1 Answer Agent, PPO achieves F1, BLEU-1, and J scores of 32.5, 24.6, and 59.4, while GRPO reaches 33.0, 24.9, and 59.9. With the full pipeline, PPO improves to 41.0, 32.9, and 57.5, and GRPO further increases performance to 45.0, 37.5, and 62.7. This demonstrates that reward-driven fine-tuning enhances answer quality beyond static retrieval. A case study is provided in Appendix~\ref{app:case_study_answer_agent}.

\paragraph{Effect of Memory Distillation}
We evaluate memory distillation by comparing Answer Agents trained with and without distillation (Figure~\ref{fig:ablation_study} (c,d)). With distillation enabled, PPO improves from 39.3, 30.9, and 57.4 to 41.0, 32.9, and 57.5 on F1, BLEU-1, and J, respectively. GRPO shows larger gains, increasing from 41.0, 34.4, and 60.1 to 45.0, 37.5, and 62.7. These results indicate that filtering irrelevant memories reduces noise and improves reasoning.

\begin{figure}[t]
\centering
\includegraphics[width=1\columnwidth]{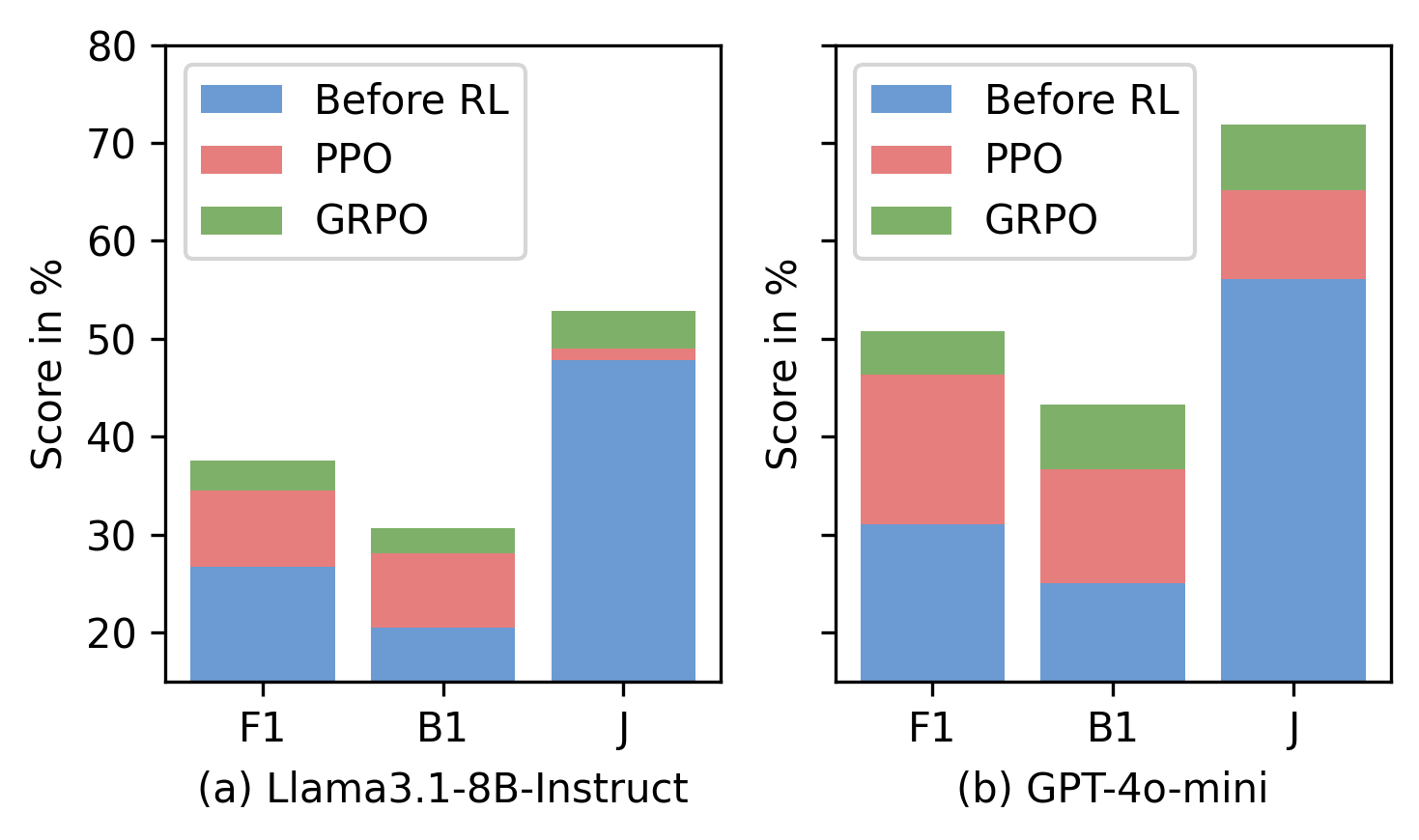} 
\caption{Performance gains of Answer Agent increase when paired with stronger Memory Managers, showing compounding benefits from higher memory quality.}
\label{answer_agent_comparison}
\end{figure}

\paragraph{RL-Fine-Tuned Answer Agent Gains More with Stronger Memory Manager}
We test whether Answer Agent gains depend on Memory Manager quality. Figure~\ref{answer_agent_comparison} compares PPO/GRPO agents with a LLaMA-3.1-8B manager versus a stronger GPT-4o-mini manager. Improvements are larger with the stronger manager (F1: +10.10 vs. +19.72; BLEU-1: +10.81 vs. +18.19; J: +5.05 vs. +15.76), showing that Memory-R1 compounds benefits and the Answer Agent scales with memory quality.

\begin{figure}[t]
\centering
\includegraphics[width=1\columnwidth]{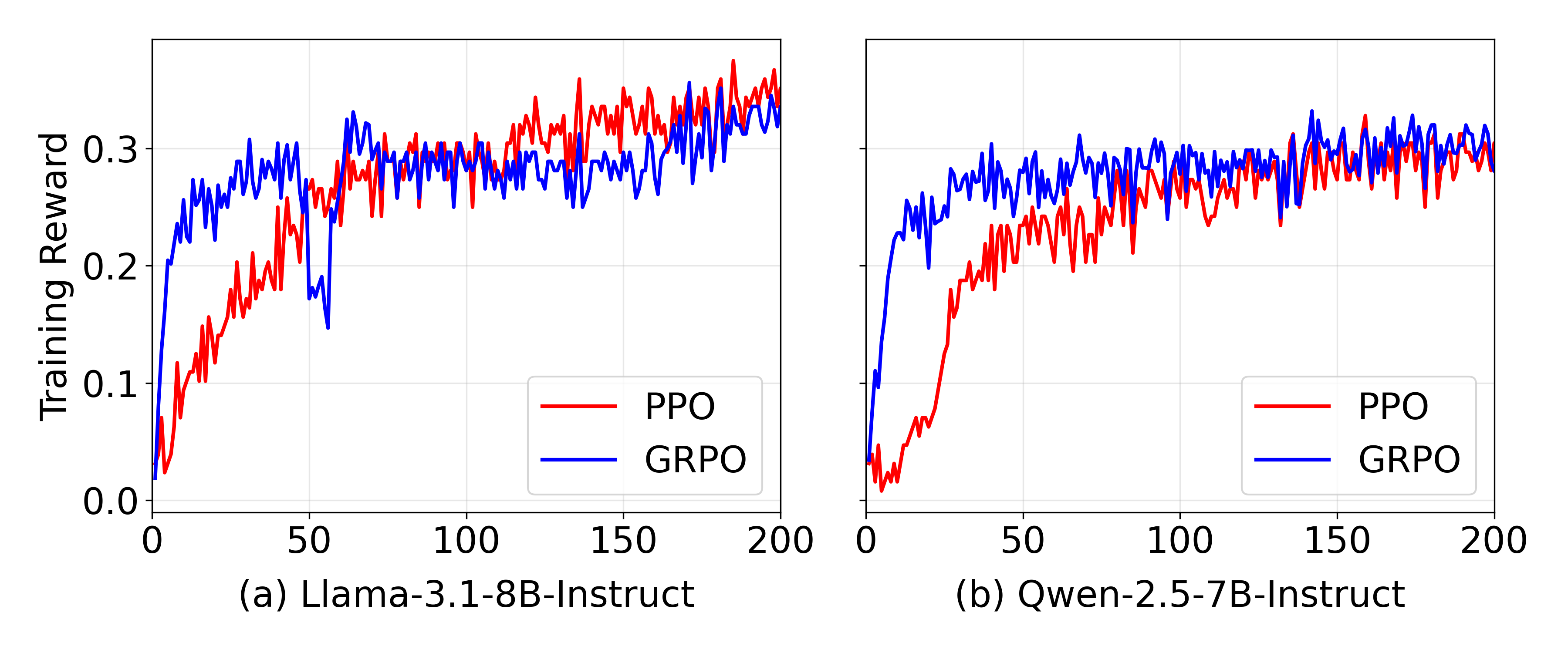} 
\caption{Training reward curves for PPO and GRPO on the Answer Agent using exact match as the reward. GRPO converges faster initially, and both reach similar final rewards.}
\label{ppo_grpo_convergence_rate}
\end{figure}

\paragraph{Comparison of RL Policies}  
We compare PPO and GRPO for training the Answer Agent, using exact match against ground‑truth answers as the reward signal. As shown in Figure~\ref{ppo_grpo_convergence_rate}, GRPO exhibits faster initial convergence, likely due to its grouped return normalization providing stronger early guidance. However, as training progresses, both methods steadily improve and ultimately reach comparable final reward levels.

\paragraph{Reward Design Analysis}
We experimented with different reward models for fine-tuning the Answer Agent. As shown in Table~\ref{tab:reward-design-ablation}, using the LLM-as-a-Judge value as reward leads to the highest J score (63.58), but performs poorly on F1 and BLEU-1. This is because the reward encourages longer, descriptive answers, which misaligns with string-overlap metrics. For example, when asked “Did John and James study together?”, the EM-based model outputs “Yes”, while the LLM-as-a-Judge–based model produces “Yes, John and James studied together, as they were part of the same online programming group, as implied by the memories above.” Although both are semantically correct, the latter is penalized under F1 and BLEU-1. This makes direct comparison with baselines difficult, since responses are no longer length-controlled. To avoid bias from relying on a single metric, we adopt the EM reward, which yields balanced improvements across all three metrics. 

\begin{table}[t]
\centering
\scalebox{0.81}{
\begin{tabular}{lccc}
\toprule
\textbf{Method} & \textbf{F1$\uparrow$} & \textbf{B1$\uparrow$} & \textbf{J$\uparrow$} \\
\midrule
PPO (J-based reward model) & 33.69 & 23.36 & \textbf{63.58} \\
PPO (EM-based reward model) & \textbf{41.05} & \textbf{32.91} & 57.54 \\
\bottomrule
\end{tabular}}
\caption{Reward Design Choice Comparison. PPO with J-based reward achieves higher J scores but lower F1 and B1 due to verbose outputs, while the EM-based reward yields balanced performance across metrics.}
\label{tab:reward-design-ablation}
\end{table}

\begin{figure}[t]
\centering
\includegraphics[width=1\columnwidth]{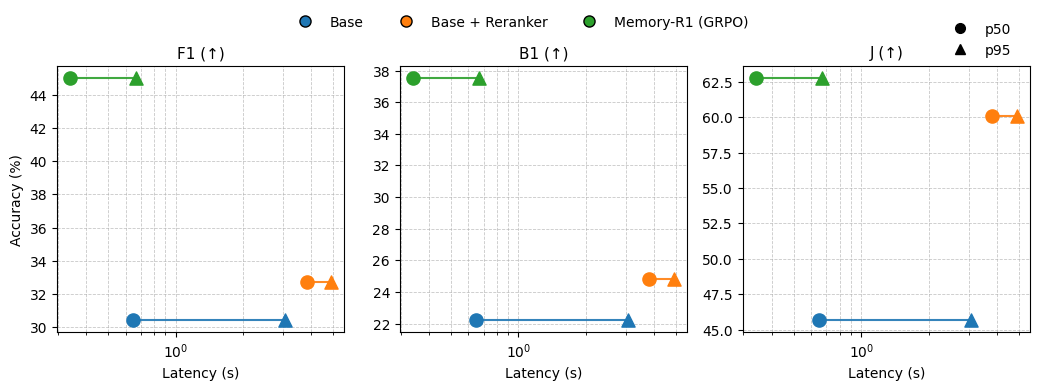} 
\caption{Accuracy and latency comparison across different inference pipelines: Base, Base + Reranker, and Memory-R1 (GRPO).}
\label{fig:acc_vs_latency}
\end{figure}

\paragraph{Comparison of Learned Memory Distillation and Reranking}
We compare learned memory distillation in Memory-R1 with reranker-based pipelines in terms of accuracy and inference latency across three settings: Base, Base + Reranker, and Memory-R1 with a GRPO-trained Answer Agent (Figure~\ref{fig:acc_vs_latency}). While reranking provides modest accuracy gains, it incurs substantial latency overhead. In contrast, Memory-R1 achieves higher accuracy with lower median and tail latency, demonstrating a more favorable accuracy–latency trade-off. Additional analyses are provided in Appendix~\ref{appendix:latency_analysis}.

\section{Conclusion}
We presented Memory-R1, a reinforcement learning framework that enables LLM-based agents to effectively manage and utilize external memory. Unlike heuristic pipelines, Memory-R1 learns memory operations as well as memory distillation and usage for answering. With only 152 training examples, it achieves state-of-the-art results on LoCoMo, scales across model sizes, and generalizes to MSC and LongMemEval without retraining. Ablation studies confirm that reinforcement learning improves every component of the system. Overall, Memory-R1 highlights reinforcement learning as a promising direction for adaptive and agentic memory in LLMs.

\section*{Limitations}
Our evaluation focuses on dialogue-centric datasets. While these benchmarks cover a wide range of reasoning types, extending Memory-R1 to multimodal data may introduce challenges beyond the scope of this work. Additionally, we train the Memory Manager and Answer Agent separately to ensure stability under sparse rewards. This separation is necessary but makes the process less straightforward. An end-to-end multi-agent reinforcement learning approach could simplify training and enable richer coordination, which we view as a promising direction for future work.

\bibliography{custom}

\clearpage
\appendix
\section{Case Study of Behavior of Agents before and after Fine-tuning}
\subsection{From In-context Memory Manager to RL fine-tuned Memory Manager}
\label{app:case_study_memory_manager}

To demonstrate how RL fine-tuning improves memory operations, we present two real representative examples. In the first case, the user initially mentions adopting a dog named Buddy, and later states that they have adopted another dog named Scout.

\noindent\textbf{Dialogue Excerpts:}
\begin{quote}
\texttt{• Audrey: That's great news! What's the pup's name?}\\
\texttt{• Andrew: It took us a while to decide, but we ended up going with 'Scout' for our pup - it seemed perfect for their adventurous spirit.}
\end{quote}

\noindent\textbf{Vanilla Memory Manager Actions (failed):}
\begin{quote}
\texttt{**Operations Issued:**}\\
\texttt{• DELETE: "Andrew adopted a new dog from a shelter and named him Buddy because he is his buddy."}\\
\texttt{• DELETE: "Andrew enjoys taking walks with his newly adopted dog, Buddy, as a way to spend quality time together and get fresh air."}\\
\texttt{• ADD: "Andrew adopted a new dog and named the pup 'Scout' because the name seemed perfect for their adventurous spirit."}\\
\end{quote}

\noindent\textbf{Memory-R1 Memory Manager Actions (successful):}
\begin{quote}
\texttt{**Operations Issued:**}\\
\texttt{• UPDATE: 
"Andrew adopted a new dog from a shelter and named him Buddy because he is his buddy." ->
"Andrew adopted a dog from a shelter and named him Buddy because he is his buddy, and later adopted another dog named Scout."}\\
\end{quote}

\noindent\textbf{Discussion:}
The vanilla LLM-based memory manager misinterpreted the second adoption as a contradiction because it saw two separate statements: “adopted a dog named Buddy” and “adopted another dog named Scout”, and assumed that the differing dog names implied the new statement replaced the old one. As a result, it deleted the first dog (“Buddy”) and added the second (“Scout”), leaving the memory fragmented and inaccurate. In contrast, the RL fine-tuned Memory Manager recognized the events as complementary and consolidated them with a single \texttt{UPDATE}, maintaining a coherent memory bank and enabling the system to answer correctly. This case illustrates how reinforcement learning enhances the Agent to understand the nuance between \texttt{ADD}, \texttt{UPDATE}, and \texttt{DELETE}, allowing the model to handle overlapping or related information more intelligently.

To further demonstrate how RL fine-tuning improves memory operations, we present another real-world example from a user-agent dialogue involving Joanna’s feelings about pets and her allergies, particularly toward turtles.

\noindent\textbf{Dialogue Excerpts:}
\begin{quote}
\texttt{• Joanna: I wish I wasn't allergic! I would get two turtles today if I could! I found out recently I'm allergic to cockroaches as well, so who knows if I'll ever get a pet.}\
\texttt{• Nate: Sorry! Maybe there are other animals you could consider! In the meantime though, I'll be sure to send you pics of my turtles so you can still watch them grow without getting too close.}
\end{quote}

\noindent\textbf{Vanilla Memory Manager Actions (failed):}
\begin{quote}
\texttt{**Operations Issued:**}\\
\texttt{• DELETE: "Joanna is allergic to most reptiles and animals with fur, which causes her face to get puffy and itchy. Despite this, she finds other ways to be happy."}\\
\texttt{• DELETE: "Joanna finds turtles peaceful and believes they bring calm and joy."}\\
\texttt{• DELETE: "Joanna thinks having pets must be a wonderful experience."}\\
\texttt{• ADD: "Joanna is allergic to turtles and cockroaches, which affects her ability to have pets."}
\end{quote}

\noindent\textbf{Memory-R1 Memory Manager Actions (successful):}
\begin{quote}
\texttt{**Operations Issued:**}\\
\texttt{• UPDATE:  "Joanna is allergic to most reptiles and animals with fur, which causes her face to get puffy and itchy. Despite this, she finds other ways to be happy." -> "Joanna is allergic to most reptiles, including turtles and cockroaches, and animals with fur, which causes her face to get puffy and itchy. Additionally, she is also allergic to turtles and cockroaches, which affects her ability to have pets."}\\
\texttt{• UPDATE: "Joanna finds turtles peaceful and believes they bring calm and joy." -> "Joanna finds turtles peaceful and believes they bring calm and joy, but she is allergic to them."}
\end{quote}

\noindent\textbf{Discussion:}
The vanilla Memory Manager misinterpreted the new allergy information in two aspects. First, it viewed the difference between “allergic to most reptiles and animals with fur” and “allergic to turtles and cockroaches” as a contradiction in allergy type—failing to recognize the latter as a more specific elaboration rather than a conflicting statement. Second, it interpreted Joanna’s expressed fondness for turtles as incompatible with her allergy to them, incorrectly assuming that emotional attachment and physical limitations cannot coexist. As a result, it issued a series of \texttt{DELETE} operations followed by a single \texttt{ADD}, discarding valuable emotional context such as Joanna’s admiration for turtles and her general enthusiasm toward pets. In contrast, the RL fine-tuned Memory Manager recognized that these pieces of information were complementary: Joanna likes turtles but cannot keep them due to her allergies. It updated the relevant memories accordingly using targeted \texttt{UPDATE} operations, preserving both factual accuracy and emotional nuance. This case demonstrates how reinforcement learning equips the model to reason about overlapping and evolving information more intelligently, favoring memory consolidation over fragmentation.

\subsection{From Vanilla LLM to Memory‑Distilled RL Answer Agent}
\label{app:case_study_answer_agent}
To illustrate how RL fine-tuned Answer Agent with \textit{Memory Distillation} improves answer accuracy, we compare the original model’s output with the RL fine-tuned model on a representative example from LoCoMo. The prompt provided to the model is shown in Figure~\ref{answer_agent_prompt}.

\noindent\textbf{Question:} \textit{Does John live close to a beach or the mountains?}

\noindent\textbf{Original Model Output (Before Fine-Tuning):}
\begin{quote}
\texttt{**Answer:** He lives near mountains.}
\end{quote}

\noindent\textbf{Memory-R1 Output (After RL Fine-Tuning with Memory Distillation):}
\begin{quote}
\texttt{**Memories selected as relevant:**}\\
\texttt{• 8:30 pm on 1 January, 2023: John has a nostalgic memory of having a film camera as a kid and taking many pictures at the beach.}\\
\texttt{• 1:24 pm on 25 May, 2023: John shared a picture of his family at the beach, expressing a sense of making a difference and commitment to continue their efforts.}\\[4pt]
\texttt{**Answer:** beach}
\end{quote}

\noindent\textbf{Discussion:} The original model consumed all retrieved memories indiscriminately and defaulted to ``mountains,'' likely influenced by irrelevant mentions of mountaineering. In contrast, Memory-R1 filtered out distractors, surfaced only beach-related memories, and generated the correct answer. This case highlights how \textit{Memory Distillation} helps the model discard noise, focus on true signals, and improve factual accuracy.

\section{Dataset Details}
\label{sec:dataset_details}

\subsection{Test Data}
\noindent\textbf{LoCoMo.}
LoCoMo~\citep{maharana2024evaluating} is a benchmark of long-term multi-session dialogues, with conversations averaging 300 turns and 9k tokens, spanning up to 35 sessions. It serves as our primary experimental dataset, on which we conduct and report detailed results.

\vspace{2mm}
\noindent\textbf{MSC.}
We further evaluate on the Multi-Session Chat (MSC) dataset~\citep{xu2021beyond}, which contains open-domain dialogues spanning multiple sessions. Following MemGPT~\citep{packer2023memgpt}, we use a modified version of MSC tailored to the memory-augmented evaluation setting, where questions depend on information distributed across earlier sessions. This dataset tests whether models can maintain continuity across temporally separated interactions.

\vspace{2mm}
\noindent\textbf{LongMemEval}
We also evaluate on LongMemEval~\citep{wu2024longmemeval}, a benchmark designed to test long-term memory capabilities of LLMs. It covers diverse tasks including factual recall, temporal reasoning, and entity tracking, with questions requiring integration of information from long and sparse contexts. LongMemEval complements LoCoMo and MSC by emphasizing broader generalization beyond dialogue-centric settings. 

\subsection{Training Data}
\label{sec:data_construction}
We construct separate training datasets for the \textit{Memory Manager} and the \textit{Answer Agent} from the LoCoMo multi-turn dialogues. The LoCoMo dataset is publicly released under the \textbf{CC BY-NC 4.0} license. We slightly modify it for dialogue segmentation to fit our reinforcement learning pipeline, 
while preserving its original license terms and using it solely for non-commercial research purposes. All other datasets used in this paper (MSC and LongMemEval) are publicly available research benchmarks and are used in accordance with their respective licenses.

\vspace{2mm}
\noindent\textbf{Memory Manager Training Data.}
For each dialogue turn $t$, GPT-4o-mini builds a temporal memory bank from the preceding 24 turns. The current turn $t$ is fused with this snapshot to form the input. Unlike supervised annotation of memory operations, we do not provide explicit labels (\texttt{ADD}, \texttt{UPDATE}, \texttt{DELETE}, \texttt{NOOP}). Instead, the Memory Manager is optimized via reinforcement learning, where the correctness of the downstream Answer Agent’s answer provides the learning signal. The full procedure is given in Algorithm~\ref{alg:data_construction_algo1}.

\vspace{2mm}
\noindent\textbf{Answer Agent Training Data.}
For each question $q$ in LoCoMo, we retrieve 60 candidate memories using retrieval-augmented search (RAG) over the temporal memory bank. The retrieved set, paired with the question and gold answer, serves as the training input for the Answer Agent, which learns to distill the relevant entries and generate concise, correct responses.

\begin{algorithm}[t]
\caption{Data Construction for Memory-R1 Training}
\label{alg:data_construction_algo1}
\begin{algorithmic}[1]
\State \textbf{Input:} LoCoMo multi-turn dialogues $\mathcal{D}$
\State \textbf{Output:} Training tuples for the Memory Manager 
$(\text{dialogue turn}, \text{temporal memory bank}, \text{QA})$

\For{each dialogue $d \in \mathcal{D}$}
    \For{each turn $t$ in $d$}
        \State Build a \textbf{temporal memory bank} using the previous 50 turns with GPT-4o-mini
        \State Combine (i) the temporal memory bank, (ii) the current turn $t$, and (iii) any QA pairs linked to $t$
        \State Store the combined package as a single training tuple
    \EndFor
\EndFor
\end{algorithmic}
\end{algorithm}

\begin{algorithm}[t]
\caption{Data Construction for Answer Agent Training}
\label{alg:answer_agent_data_construction}
\begin{algorithmic}[1]
\State \textbf{Input:} LoCoMo multi-turn dialogues $\mathcal{D}$, trained Memory Manager
\State \textbf{Output:} Training tuples for the Answer Agent 
$(\text{question}, \text{retrieved memories}, \text{gold answer})$

\For{each dialogue $d \in \mathcal{D}$}
\State Use the Memory Manager to maintain an up-to-date memory bank across turns
\EndFor
\For{each question $q$ in $d$}
    \State Use the question $q$ as a query to retrieve the top 30 most relevant candidate memories for each participant from the memory bank
    \State Pair (i) the question $q$, (ii) the 60 retrieved memories, and (iii) the gold answer $a_{\text{gold}}$
    \State Store the triplet as a single training tuple for Answer Agent fine-tuning
\EndFor
\end{algorithmic}
\end{algorithm}

\section{Prompts}
\label{appendix_prompt}
In developing our Memory Manager Prompt, answer generation agent prompt, and LLM-as-a-Judge prompt, we adapt elements from the prompt released by prior work~\citep{packer2023memgpt, chhikara2025mem0}

\subsection{Memory Manager Prompt}

For training the Memory Manager, we use a detailed prompt that instructs the model how to perform four memory operations: \texttt{ADD}, \texttt{UPDATE}, \texttt{DELETE}, and \texttt{NOOP}. The full prompt spans multiple figures for readability.

\vspace{2mm}
\begin{figure*}[htbp]
\centering
\begin{minipage}{0.9\textwidth}
\scriptsize

\begin{tcolorbox}[promptstyle={Memory Manager Prompt (Part 1): Overview and ADD/UPDATE Instruction}]
\begin{verbatim}
You are a smart memory manager which controls the memory of a system.
You can perform four operations: (1) add into the memory, (2) update the 
memory, (3) delete from the memory, and (4) no change.

Based on the above four operations, the memory will change.

Compare newly retrieved facts with the existing memory. For each new fact,
decide whether to:
- ADD: Add it to the memory as a new element
- UPDATE: Update an existing memory element
- DELETE: Delete an existing memory element
- NONE: Make no change (if the fact is already present or irrelevant)

1. **Add**: If the retrieved facts contain new information not present 
in the memory, then you have to add it by generating a new ID in the id field.

- Example:
    Old Memory:
        [
            {"id" : "0", "text" : "User is a software engineer"}
        ]
    Retrieved facts: ["Name is John"]

    New Memory:
        {
            "memory" : [
                {"id" : "0", "text" : "User is a software engineer", "event" : "NONE"},
                {"id" : "1", "text" : "Name is John", "event" : "ADD"}
            ]
        }
2. **Update**: If the retrieved facts contain information that is already 
present in the memory but the information is totally different, then 
you have to update it.

If the retrieved fact contains information that conveys the same thing as 
the memory, keep the version with more detail.

Example (a) – if the memory contains "User likes to play cricket" and the 
retrieved fact is "Loves to play cricket with friends", then update the 
memory with the retrieved fact.

Example (b) – if the memory contains "Likes cheese pizza" and the 
retrieved fact is "Loves cheese pizza", then do NOT update it because they 
convey the same information.

Important: When updating, keep the same ID and preserve old_memory.

- Example:
    Old Memory:
        [
            {"id" : "0", "text" : "I really like cheese pizza"},
            {"id" : "2", "text" : "User likes to play cricket"}
        ]
    Retrieved facts: ["Loves chicken pizza", "Loves to play cricket with friends"]

    New Memory:
        {
        "memory" : [
            {"id" : "0", "text" : "Loves cheese and chicken pizza", "event" : "UPDATE",
             "old_memory" : "I really like cheese pizza"},
            {"id" : "2", "text" : "Loves to play cricket with friends", "event" : "UPDATE",
             "old_memory" : "User likes to play cricket"}
        ]
        }
\end{verbatim}
\end{tcolorbox}

\caption{Memory Manager Prompt (Part 1): Overview and ADD/UPDATE operation instruction.}
\label{fig:mm_prompt_part1}
\end{minipage}
\end{figure*}

\vspace{2mm}
\begin{figure*}[t]
\centering
\begin{minipage}{0.9\textwidth}
\scriptsize

\begin{tcolorbox}[promptstyle={Memory Manager Prompt (Part 2): DELETE/NO\_OPERATION Instructions}]
\begin{verbatim}
3. **Delete**: If the retrieved facts contain information that contradicts 
the memory, delete it. When deleting, return the same IDs — do not generate new IDs.

- Example:
    Old Memory:
        [
            {"id" : "1", "text" : "Loves cheese pizza"}
        ]
    Retrieved facts: ["Dislikes cheese pizza"]

    New Memory:
        {
        "memory" : [
            {"id" : "1", "text" : "Loves cheese pizza", "event" : "DELETE"}
        ]
        }

4. **No Change**: If the retrieved facts are already present, make no change.

- Example:
    Old Memory:
        [
            {"id" : "0", "text" : "Name is John"}
        ]
    Retrieved facts: ["Name is John"]

    New Memory:
        {
        "memory" : [
            {"id" : "0", "text" : "Name is John", "event" : "NONE"}
        ]
        }
\end{verbatim}
\end{tcolorbox}

\caption{Memory Manager Prompt (Part 2): DELETE/NO\_OPERATION instructions.}
\label{fig:mm_prompt_part2}
\end{minipage}
\end{figure*}

\subsection{Answer Agent Prompt}

We provide the full prompt used to instruct the Answer Agent in our case study. This prompt defines the reasoning process, memory selection criteria, and formatting requirements for the model’s responses. Figure~\ref{answer_agent_prompt} shows the complete instructions, context, and representative retrieved memories.
\vspace{2mm}
\begin{figure*}[t]
\centering
\begin{minipage}{0.9\textwidth}
\scriptsize

\begin{tcolorbox}[promptstyle={Full Prompt and Retrieved Memories}]
\begin{verbatim}
You are an intelligent memory assistant tasked with retrieving
accurate information from conversation memories.

# CONTEXT:
You have access to memories from two speakers in a conversation.
These memories contain timestamped information that may be relevant
to answering the question.

# INSTRUCTIONS:
1. Carefully analyze all provided memories from both speakers
2. Pay special attention to the timestamps to determine the answer
3. If the question asks about a specific event or fact, look for direct evidence
4. If the memories contain contradictory information, prioritize the most recent memory
5. If there is a question about time references (like "last year", "two months ago"), 
   calculate the actual date based on the memory timestamp.
6. Always convert relative time references to specific dates, months, or years.
7. Focus only on the content of the memories. Do not confuse character names
8. The answer should be less than 5-6 words.
9. IMPORTANT: Select memories you found that are useful for answering the questions, 
and output it before you answer questions.
10. IMPORTANT: Output the final answer after **Answer:**

# APPROACH (Think step by step):
1. Examine all relevant memories
2. Examine the timestamps carefully
3. Look for explicit mentions that answer the question
4. Convert relative references if needed
5. Formulate a concise answer
6. Double-check the answer correctness
7. Ensure the final answer is specific
8. First output the memories that you found are important before you answer questions

Memories for user John:
- 7:20 pm on 16 June, 2023: John has a special memory of a vacation to California where he experienced a 
gorgeous sunset and an enjoyable night strolling the shore, creating meaningful memories with loved ones.
- 6:13 pm on 10 April, 2023: John explored the coast in the Pacific Northwest and visited some national
parks, finding the beauty of nature absolutely breathtaking.
- 3:14 pm on 13 August, 2023: John enjoys spending time outdoors with his family, including activities 
such as hiking, hanging out at the park, and having picnics. He also values indoor family activities like 
playing board games and having movie nights at home.
... (In total 30 most relevant memories from John's Memory Bank are provided) ...

Memories for user Maria:
- 6:29 pm on 7 July, 2023: John experienced a severe flood in his old area last week, which caused 
significant damage to homes due to poor infrastructure.
- 1:24 pm on 25 May, 2023: Maria appreciates the beauty of small, meaningful moments in life, as reflected 
in her reaction to a family beach photo shared by John.
- 3:14 pm on 13 August, 2023: Maria appreciates family bonding and is interested in the activities that 
John and his family enjoy doing together.
... (In total 30 most relevant memories from Maria's Memory Bank are provided) ...

Question: Does John live close to a beach or the mountains?
\end{verbatim}
\end{tcolorbox}

\caption{Prompt and retrieved memories used in the case study, showing all instructions, context, and memory entries provided to the model.}
\label{answer_agent_prompt}
\end{minipage}
\end{figure*}

\subsection{LLM-as-a-Judge (J) Prompt}

For evaluating the correctness of generated answers, we employ an LLM-as-a-Judge prompt. The judge model is asked to label each answer as \texttt{CORRECT} or \texttt{WRONG} based on comparison with the gold answer. The complete prompt template is shown in Figure~\ref{llm_judge_prompt}.

\vspace{2mm}
\begin{figure*}[t]
\centering
\begin{minipage}{0.9\textwidth}
\scriptsize

\begin{tcolorbox}[promptstyle={LLM-as-a-Judge Prompt Template}]
\begin{verbatim}
Your task is to label an answer to a question as 'CORRECT' or 'WRONG'.
You will be given the following data:
    (1) a question (posed by one user to another user),
    (2) a 'gold' (ground truth) answer,
    (3) a generated answer,
which you will score as CORRECT or WRONG.

The point of the question is to ask about something one user should know about the other user based on their 
prior conversations.

The gold answer will usually be a concise and short answer that includes the referenced topic, for example:
Question: Do you remember what I got the last time I went to Hawaii?
Gold answer: A shell necklace

The generated answer might be longer, but you should be generous with your grading — as long as it touches 
on the same topic as the gold answer, it should be counted as CORRECT.

For time-related questions, the gold answer will be a specific date, month, or year. The generated answer 
might include relative references (e.g., "last Tuesday"), but you should be generous — if it refers to 
the same time period as the gold answer, mark it CORRECT, even if the format differs (e.g., "May 7th" vs. 
"7 May").

Now it's time for the real question:
Question: {question}
Gold answer: {gold_answer}
Generated answer: {generated_answer}

First, provide a short (one sentence) explanation of your reasoning, then finish with CORRECT or WRONG. 
Do NOT include both CORRECT and WRONG in your response, or it will break the evaluation script.

Return the label in JSON format with the key as "label".
\end{verbatim}
\end{tcolorbox}

\caption{LLM-as-a-Judge prompt used to evaluate model answers. The judge model labels each generated answer as CORRECT or WRONG based on comparison with the gold answer, with explicit instructions for handling time references and topic matching.}
\label{llm_judge_prompt}
\end{minipage}
\end{figure*}

\section{Implementation Details}
\label{appendix:implementation-details}

We fine-tune \textsc{Memory-R1} on LLaMA-3.1-8B-Instruct and Qwen-2.5-3B, 7B, and 14B-Instruct models to evaluate robustness across architectures. Experiments are primarily conducted on 4 NVIDIA H100 GPUs (80GB each), except for Qwen-2.5-14B, which requires 8 GPUs. The total batch size is 128 with a micro-batch size of 2 per GPU. The maximum prompt and response lengths are set to 4096 and 2048 tokens, respectively.

Prompts for memory operations and memory-augmented answer generation are adapted from~\citet{chhikara2025mem0}. Reinforcement learning fine-tuning is performed using PPO and GRPO within the VERL framework~\citep{sheng2025hybridflow}. For PPO, actor and critic networks are jointly trained with learning rates of $1\times10^{-6}$ and $1\times10^{-5}$, respectively, using a constant warmup schedule. GRPO updates only the actor via grouped return normalization.

During RL training, we use a decoding temperature of $\tau=1.0$ to encourage exploration and collect diverse reward signals, which helps stabilize policy learning. For validation and testing, greedy decoding ($\tau=0$) is applied to ensure deterministic outputs and consistent metric evaluation.

\section{Alogirthm}
The overall Memory-R1 pipeline contains two complementary procedures, outlined in Algorithm ~\ref{alg:memory_construction} and Algorithm~\ref{alg:answer_generation}. Algorithm~\ref{alg:memory_construction} (Memory Bank Construction) governs how the system incrementally builds and refines the external memory bank as new dialogue turns arrive. For each dialogue input, an LLM extracts key information, retrieves semantically related entries from the memory bank via retrieval-augmented generation (RAG), and invokes the RL fine-tuned Memory Manager to classify the update action as one of \{\texttt{ADD}, \texttt{UPDATE}, \texttt{DELETE}, \texttt{NOOP}\}. Depending on the chosen action, the memory store is updated accordingly—either inserting a new entry, merging information into an existing one, pruning contradictory content, or leaving the memory unchanged.

Algorithm~\ref{alg:answer_generation} (Memory-augmented Answer Generation) describes how the system leverages the constructed memory bank to generate answers. Given an incoming question, the model retrieves the top‑k relevant memory candidates, concatenates them with the question to form a memory-augmented prompt, and applies the Answer Agent’s Memory Distillation policy to filter for the most relevant facts. The distilled memory context, along with the query, is then passed to the Answer Agent to produce the final response, which is added to the answer set. Together, these algorithms enable Memory-R1 to jointly manage memory and generate memory augmented answers.

Training in Memory-R1 is performed in two stages, with the Memory Manager and Answer Agent optimized separately. When training the Memory Manager, the Answer Agent is frozen and used only to provide outcome-based rewards: the Manager’s operations are reinforced if the resulting memory state improves the Answer Agent’s ability to answer correctly. Conversely, when training the Answer Agent, the Memory Manager is fixed to ensure a stable memory input. Algorithm~\ref{alg:memory_manager_rl_pipeline} illustrates this process for the Memory Manager, where dialogue turns are processed sequentially, candidate operations are sampled, the memory bank is updated, and policy gradients (via PPO or GRPO) are applied based on downstream answer correctness. This decoupled setup avoids attribution ambiguity while still allowing both components to co-adapt over alternating training phases.

\begin{algorithm}[t]
\small
\caption{Memory Bank Construction via Memory Manager}
\label{alg:memory_construction}
\begin{algorithmic}[1]

\State \textbf{Input:} Multi-turn dialogue $D = \{ t_1, t_2, \dots, t_n \}$; Initial empty memory bank $M$
\State \textbf{Output:} Updated memory bank $M$

\Procedure{ConstructMemoryBank}{$D, M$}

    \For{each dialogue turn $t_i \in D$}
        \State Extract key info: $f_i \gets \text{LLMExtract}(t_i)$
        \State Retrieve memories:$M_{old} \gets \text{TopK}(f_i, M)$
        \State Determine operation: 
        \State $o_i \gets \text{MemoryManager}(f_i, M_{old})$ where $o_i \in \{\texttt{ADD}, \texttt{UPDATE}, \texttt{DELETE}, \texttt{NOOP}\}$
        \If{$o_i = \texttt{ADD}$}
            \State $M \gets M \cup \{ f_i \}$
        \ElsIf{$o_i = \texttt{UPDATE}$}
            \State $M_{tmp} \gets \text{Merge}(M_{old}, f_i)$
            \State $M \gets M \setminus M_{old} \cup M_{tmp}$
        \ElsIf{$o_i = \texttt{DELETE}$}
            \State $M \gets M \setminus M_{old}$
        \ElsIf{$o_i = \texttt{NOOP}$}
            \State $M \gets M$
        \EndIf

    \EndFor

    \State \textbf{return} $M$

\EndProcedure

\end{algorithmic}
\end{algorithm}

\begin{algorithm}[t]
\small
\caption{Memory-augmented Generation via Answer Agent}
\label{alg:answer_generation}
\begin{algorithmic}[1]

\State \textbf{Input:} Question set $Q = \{q_1, q_2, \ldots, q_m\}$; Memory bank $M$; Generation instruction text $t$
\State \textbf{Output:} Answer set $\hat{A}$

\Procedure{GenerateAnswers}{$Q, M, t$}
    \State $\hat{A} \gets \{\}$
    \For{each question $q_i \in Q$}
        \State $ M_{ret} \gets \text{TopK}(q_i, M)$
        \State $p_i \gets \text{Concat}(t, q_i,M_{ret})$  \Comment{$p_i$ is the memory augmented prompt}
        \State $ M_{distill}, \hat{a}_i \gets \text{AnswerAgent}(p_i)$ 
        \State $\hat{A} \gets \hat{A} \cup \{\hat{a}_i\}$
    \EndFor
    \State \textbf{return} $\hat{A}$
\EndProcedure

\end{algorithmic}
\end{algorithm}

\begin{algorithm}[t]
\small
\caption{Memory-R1 Pipeline for Memory Manager}
\label{alg:memory_manager_rl_pipeline}
\begin{algorithmic}[1]

\State \textbf{Input:} Dataset $D$ of tuples: dialogue turns $ds$,  question-answer pairs $(q_i, a_i)$; Temp memory bank $M$;
Memory Manager LLM $\mathcal{L}_{m}$; Answer LLM $\mathcal{L}_{a}$; Reward Function $\mathcal{F}$; Generation instruction text $t$

\State \textbf{Output:} Fine-tuned Memory Manager LLM $\mathcal{L}_{m}$
\Procedure{TrainMemoryManager}{$D, \mathcal{L}_{m}, \mathcal{L}_{a}, \mathcal{F}$}
    \For{each tuple $(ds, q_i, a_i) \in D$}
        \State $ M \gets \{\} $
        \For{$d_i \in ds$}
        \State Facts Extraction: $f_i \gets \text{LLMExtract}(d_i)$
        \State Memory Retrieval: $M_{ret} \gets \text{TopK}(f_i, M)$
        \State Determine operation: $o_i \sim \mathcal{L}_{m}(f_i, M_{ret})$ 
        \If{$o_i = \texttt{ADD}$}
            \State $M \gets M \cup \{ f_i \}$
        \ElsIf{$o_i = \texttt{UPDATE}$}
            \State $M_{tmp} \gets \text{Merge}(M_{ret}, f_i)$
            \State $M \gets M \cup M_{tmp}$
        \ElsIf{$o_i = \texttt{DELETE}$}
            \State $M \gets M \setminus M_{ret}$
        \ElsIf{$o_i = \texttt{NOOP}$}
            \State $M \gets M$
        \EndIf
        \EndFor
        \State Get Context: $C_{ret} \gets \text{TopK}(q_i, M)$ 
        \State Update Prompt: $p_i \gets \text{Concat}(t, q_i,C_{ret})$
        \State Get Response: $r_i \sim \mathcal{L}_{a}(p_i)$
        \State Policy Update: $\mathcal{L}_{m} \gets \text{RL}_{step}(\mathcal{L}_{m}, \mathcal{F}, a_i, r_i)$,
        \State where RL $ \in \{PPO, GRPO\}$ 

    \EndFor
    \State \textbf{return} $\mathcal{L}_{m}$
\EndProcedure

\end{algorithmic}
\end{algorithm}

\section{Extended Results and Type-Level Analysis}
\label{appendix:extended_results}

Tables~\ref{tab:qwen-all-model-extended-results} and \ref{tab:longmemeval-extended-results} provide detailed type-level evaluation on the LoCoMo and LongMemEval benchmarks. On LoCoMo (Table~\ref{tab:qwen-all-model-extended-results}), Memory-R1 achieves consistent improvements across all reasoning types, with the largest gains on multi-hop and temporal questions, confirming its ability to maintain and integrate long-range information.

On LongMemEval (Table~\ref{tab:longmemeval-extended-results}), improvements are most pronounced in multi-session scenarios where continuity across temporally distant interactions is critical. Memory-R1 shows substantial gains on tasks requiring factual recall (SSU) and temporal reasoning (TR), while also yielding steady improvements in knowledge update (KU) and open-domain QA. Across reasoning types, GRPO generally outperforms PPO, particularly in scenarios involving reasoning over multiple or noisy memory entries.

In addition to type-level analysis, Table~\ref{tab:longmemeval-overall-results} reports overall performance on the LongMemEval benchmark, including all baseline methods as well as Memory-R1 variants. Importantly, Memory-R1 is fine-tuned only on the LoCoMo dataset and evaluated on LongMemEval without any additional training. Despite this zero-shot transfer setting, Memory-R1-GRPO outperforms all baseline systems across both LLaMA-3.1-8B and Qwen-2.5-7B backbones. Together, these results complement the main findings in Section~\ref{section:experiments}, further reinforcing that Memory-R1 generalizes robustly across reasoning types, model families, and benchmark tasks.

\begin{table*}[t]
\centering
\scalebox{0.65}{
\begin{tabular}{l|l|ccc|ccc|ccc|ccc|ccc}
\toprule
\textbf{Model} & \textbf{Method} & 
\multicolumn{3}{c|}{\textbf{Single Hop}} & 
\multicolumn{3}{c|}{\textbf{Multi-Hop}} & 
\multicolumn{3}{c|}{\textbf{Open Domain}} & 
\multicolumn{3}{c|}{\textbf{Temporal}} & 
\multicolumn{3}{c}{\textbf{Overall}} \\
& & F1$\uparrow$ & B1$\uparrow$ & J$\uparrow$ 
& F1$\uparrow$ & B1$\uparrow$ & J$\uparrow$ 
& F1$\uparrow$ & B1$\uparrow$ & J$\uparrow$ 
& F1$\uparrow$ & B1$\uparrow$ & J$\uparrow$ 
& F1$\uparrow$ & B1$\uparrow$ & J$\uparrow$ \\
\midrule

\multirow{3}{*}{\rotatebox{0}{%
    \begin{tabular}{c}
        \textbf{Qwen} \\
        \textbf{2.5-3B}
    \end{tabular}}}
& BASE          & 19.82 & 15.78 & 46.44 & 11.57 & 10.22 & 24.10 & 25.37 & 20.04 & 45.67 & 28.94 & 24.19 & 29.46 & 24.18 & 19.46 & 41.24 \\
& PPO            & 28.60 & 19.02 & \textbf{50.63} & \textbf{26.57} & \textbf{22.72} & 40.96 & 41.06 & 35.60 & 58.73 & 43.92 & 29.73 & 50.00 & 38.42 & 30.59 & 54.40 \\
& GRPO          & \textbf{30.10} & \textbf{20.00} & 49.37 & 25.29 & 22.69 & \textbf{44.58} & \textbf{43.01} & \textbf{37.39} & \textbf{65.06} & \textbf{47.69} & \textbf{33.36} & \textbf{50.00} & \textbf{40.45} & \textbf{32.48} & \textbf{57.92} \\

\cmidrule(lr){1-17}

\multirow{3}{*}{\rotatebox{0}{%
    \begin{tabular}{c}
        \textbf{Qwen} \\
        \textbf{2.5-7B}
    \end{tabular}}}
& BASE         & 23.61 & 17.78 & 60.67 & 17.86 & 14.40 & 43.37 & 31.39 & 24.34 & \textbf{65.75} & 32.66 & 27.51 & 40.31 & 29.36 & 23.14 & \textbf{58.38} \\
& PPO            & \textbf{34.92} & \textbf{25.69} & \textbf{59.00} & 25.30 & \textbf{22.66} & 42.17 & 43.51 & 38.00 & 65.34 & 42.52 & 30.10 & \textbf{41.86} & 40.59 & 33.21 & 58.07 \\
& GRPO          & 33.98 & 25.54 & 58.16 & \textbf{25.50} & 21.63 & \textbf{46.99} & \textbf{44.72} & \textbf{48.99} & 64.65 &\textbf{ 43.54} & \textbf{35.52} & 39.92 & \textbf{41.31} & \textbf{34.74} & 57.46 \\
\cmidrule(lr){1-17}

\multirow{3}{*}{\rotatebox{0}{%
    \begin{tabular}{c}
        \textbf{Qwen-} \\
        \textbf{2.5-14B}
    \end{tabular}}}
& BASE         & 34.60 & 26.82 & 55.23 & 24.24 & 21.45 & 38.55 & 39.79 & 34.53 & 56.95 & 34.98 & 29.39 & 33.33 & 36.91 & 31.28 & 50.80 \\
& PPO            & 37.59 & \textbf{31.92} & 63.18 & \textbf{28.21} & \textbf{24.57} & \textbf{50.60} & \textbf{48.46} & \textbf{42.44} & \textbf{72.76} & 43.78 & 34.73 & 50.78 & \textbf{44.26} & \textbf{37.86} & \textbf{65.26} \\
& GRPO          & \textbf{38.32 }& 30.64 & \textbf{63.18} & 22.71 & 20.40 & 42.17 & 46.70 & 41.70 & 67.13 & \textbf{50.50} & \textbf{36.60} & \textbf{60.47} & 44.40 & 37.32 & 63.50 \\
\bottomrule
\end{tabular}}
\caption{Extended evaluation of Memory-R1 with Qwen-2.5 model family as backbones on the LoCoMo benchmark. Results are reported across question types (Single-Hop, Multi-Hop, Open-Domain, Temporal) and overall performance. Best scores are highlighted in bold.}
\label{tab:qwen-all-model-extended-results}
\end{table*}

\begin{table*}[t]
\centering
\scalebox{0.85}{
\begin{tabular}{l|ccc|ccc}
\toprule
\textbf{Task} & 
\multicolumn{3}{c|}{\textbf{LLaMA-3.1-8B}} & 
\multicolumn{3}{c}{\textbf{Qwen-2.5-7B}} \\
& BASE & PPO & GRPO & BASE & PPO & GRPO \\
\midrule
SSU (F1/B1/J)   & 61.9/53.2/80.0 & \textbf{78.9}/75.6/\textbf{87.1} & 76.0/70.3/\textbf{87.1} 
                    & 64.4/54.6/90.0 & 70.8/65.5/80.0 & \textbf{80.9}/\textbf{76.3}/\textbf{91.4} \\
SSP (F1/B1/J)   & 7.4/0.1/46.7   & 9.6/1.6/50.0   & \textbf{11.5}/\textbf{3.8}/\textbf{63.3}
                    & 13.9/\textbf{2.5}/53.3  & \textbf{14.9}/2.2/\textbf{66.7}  & 12.6/2.0/\textbf{66.7} \\
OD (F1/B1/J)   & 17.9/16.6/19.6 & 30.6/24.6/33.9 & \textbf{31.2}/25.3/\textbf{33.9}
                    & 14.1/14.4/16.1 & 23.5/21.1/19.6 & \textbf{26.8}/\textbf{23.0}/26.8 \\
MS (F1/B1/J)        & 20.8/19.6/33.1 & 43.1/43.6/54.1 & \textbf{50.0}/\textbf{48.1}/\textbf{57.9}
                    & 30.2/26.9/54.9 & 32.4/35.1/36.1 & \textbf{51.7}/\textbf{48.5}/\textbf{63.2} \\
KU (F1/B1/J)   & 36.0/27.9/51.3 & \textbf{46.4}/{43.1}/{55.1} & 38.5/35.5/52.6
                    & 40.5/33.5/59.0 & 52.3/48.2/65.4 & \textbf{54.4}/\textbf{51.3}/\textbf{65.4} \\
TR (F1/B1/J)   & 34.0/23.1/42.1 & 37.0/29.2/\textbf{49.6} & \textbf{41.5}/\textbf{30.3}/45.1
                    & 36.5/24.5/\textbf{44.4} & \textbf{38.1}/\textbf{26.3}/38.4 & 35.1/25.8/41.4 \\
\midrule
O (F1/B1/J)   & 31.3/25.0/44.2 & 43.6/\textbf{39.5}/55.2 & \textbf{45.2}/39.3/\textbf{55.4}
                    & 35.5/28.3/53.2 & 40.3/35.5/47.4 & \textbf{46.7}/\textbf{41.1}/\textbf{57.8} \\
\bottomrule
\end{tabular}}
\caption{Extended evaluation of Memory-R1 on the LongMemEval benchmark using LLaMA and Qwen backbones. 
Each cell shows F1/B1/J for a given model-method combination, reported with one decimal precision. Task types are abbreviated as: SSU = Single-Session-User, SSP = Single-Session-Preference, OD = Open Domain, MS = Multi-Session, KU = Knowledge Update, TR = Temporal Reasoning, and O = Overall. The best value for each metric (F1, B1, J) within a task row is highlighted in bold.}
\label{tab:longmemeval-extended-results}
\end{table*}

\begin{table}[t]
\centering
\scalebox{0.55}{
\begin{tabular}{l|l|ccc}
\toprule
\textbf{Base Model} & \textbf{Method} & \textbf{Overall F1} $\uparrow$ & \textbf{Overall B1} $\uparrow$ & \textbf{Overall J} $\uparrow$ \\
\midrule
\multirow{7}{*}{LLaMA-3.1-8B} 
& LoCoMo (RAG)  & 20.55 & 15.17 & 21.00 \\
& A-Mem          & 38.36 & 33.30 & 54.20 \\
& Mem0           & 31.41 & 21.69 & 41.20 \\
& Memory-SFT     & 43.89 & 36.72 & 54.80 \\
\cmidrule{2-5}
& \cellcolor{gray!15}Memory-R1-PPO  & \cellcolor{gray!15}43.60 & \cellcolor{gray!15}\textbf{39.50} & \cellcolor{gray!15}55.20 \\
& \cellcolor{gray!15}Memory-R1-GRPO & \cellcolor{gray!15}\textbf{45.20} & \cellcolor{gray!15}39.30 & \cellcolor{gray!15}\textbf{55.40} \\

\midrule

\multirow{7}{*}{Qwen-2.5-7B} 
& LoCoMo (RAG)  & 18.27 & 14.57 & 22.20 \\
& A-Mem          & 41.55 & 36.58 & 54.80 \\
& Mem0           & 38.44 & 34.53 & 46.80 \\
& Memory-SFT     & 43.16 & 35.04 & 54.80 \\
\cmidrule{2-5}
& \cellcolor{gray!15}Memory-R1-PPO  & \cellcolor{gray!15}40.30 & \cellcolor{gray!15}35.50 & \cellcolor{gray!15}47.40 \\
& \cellcolor{gray!15}Memory-R1-GRPO & \cellcolor{gray!15}\textbf{46.70} & \cellcolor{gray!15}\textbf{41.10} & \cellcolor{gray!15}\textbf{57.80} \\
\bottomrule
\end{tabular}}
\caption{Overall results on the LongMemEval benchmark. We report the mean scores across all six evaluation dimensions. The best results are marked in \textbf{bold}.}
\label{tab:longmemeval-overall-results}
\end{table}

\section{Latency Analysis}
\label{appendix:latency_analysis}

We provide a detailed latency analysis to better understand the efficiency characteristics of Memory-R1 and its individual components. All latency results are reported using median (p50) and tail (p95) inference time, measured across three components of the pipeline: the Memory Manager, Memory Search, and the Answer Agent. We compare the base model, PPO-trained variants, and GRPO-trained variants on both LLaMA-3.1-8B and Qwen-2.5-7B backbones.

\paragraph{Overall Trends}
Across both model families, Memory-R1 does not introduce prohibitive latency overhead despite incorporating explicit memory management and reasoning components. In many cases, GRPO-trained variants achieve lower tail latency than both the base model and PPO variants, indicating that reinforcement learning can improve not only accuracy but also inference efficiency.

\paragraph{Memory Manager Latency}
For the Memory Manager component, latency remains relatively stable across Base, PPO, and GRPO variants. On LLaMA-3.1-8B, median latency ranges narrowly between 1.98\,s and 2.17\,s, with p95 latency around 3.4-3.6\,s. Similar behavior is observed on Qwen-2.5-7B, where p50 latency stays below 1.4\,s across all variants. These results suggest that RL fine-tuning does not materially increase the computational cost of memory operation selection.

\paragraph{Memory Search Latency}
Memory Search exhibits consistently low latency across all settings. On both backbones, median latency remains below 0.35\,s, and p95 latency remains under 0.65\,s. Differences between Base, PPO, and GRPO variants are minimal, indicating that improvements in downstream accuracy are not driven by more expensive retrieval operations.

\paragraph{Answer Agent Latency}
The Answer Agent shows the most pronounced latency differences across methods. On LLaMA-3.1-8B, the GRPO-trained Answer Agent achieves substantially lower median and tail latency, with p50 and p95 of 0.34\,s and 0.67\,s, compared to 0.65\,s and 3.07\,s for the base model and 0.91\,s and 4.67\,s for PPO. A similar pattern holds on Qwen-2.5-7B, where GRPO reduces p95 latency to 0.83\,s, compared to 1.06\,s for the base model and 2.60\,s for PPO. This reduction suggests that GRPO encourages more concise and efficient reasoning paths during answer generation.

\begin{figure}[t]
\centering
\includegraphics[width=1\columnwidth]{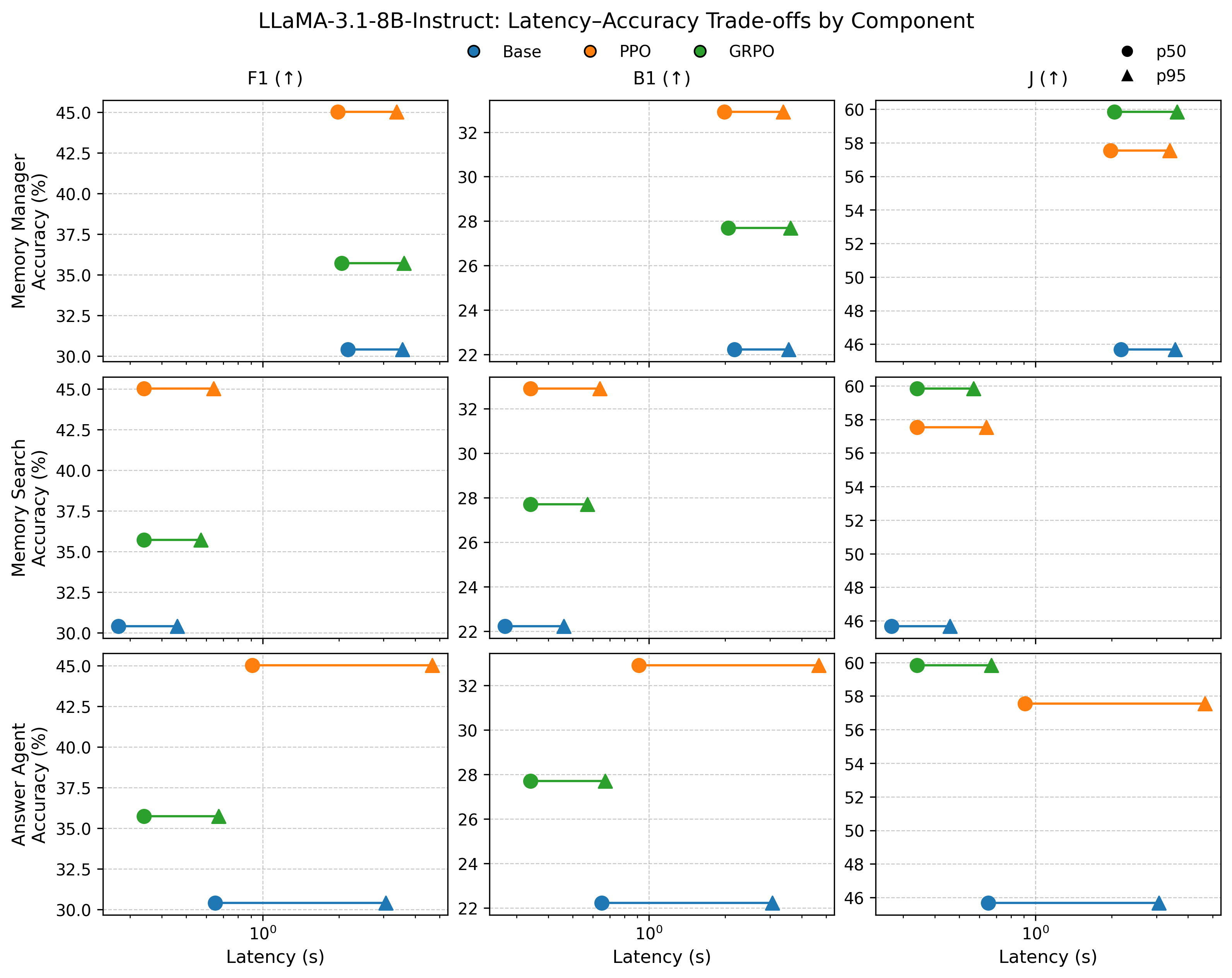} 
\caption{Latency-accuracy comparison across pipeline components on LLaMA-3.1-8B-Instruct. Points show median (p50) and tail (p95) latency versus accuracy (F1, BLEU-1, and LLM-as-a-Judge) for the base model and RL-trained variants (PPO, GRPO).}
\label{fig:latency_accuracy_llama}
\end{figure}

\begin{figure}[t]
\centering
\includegraphics[width=1\columnwidth]{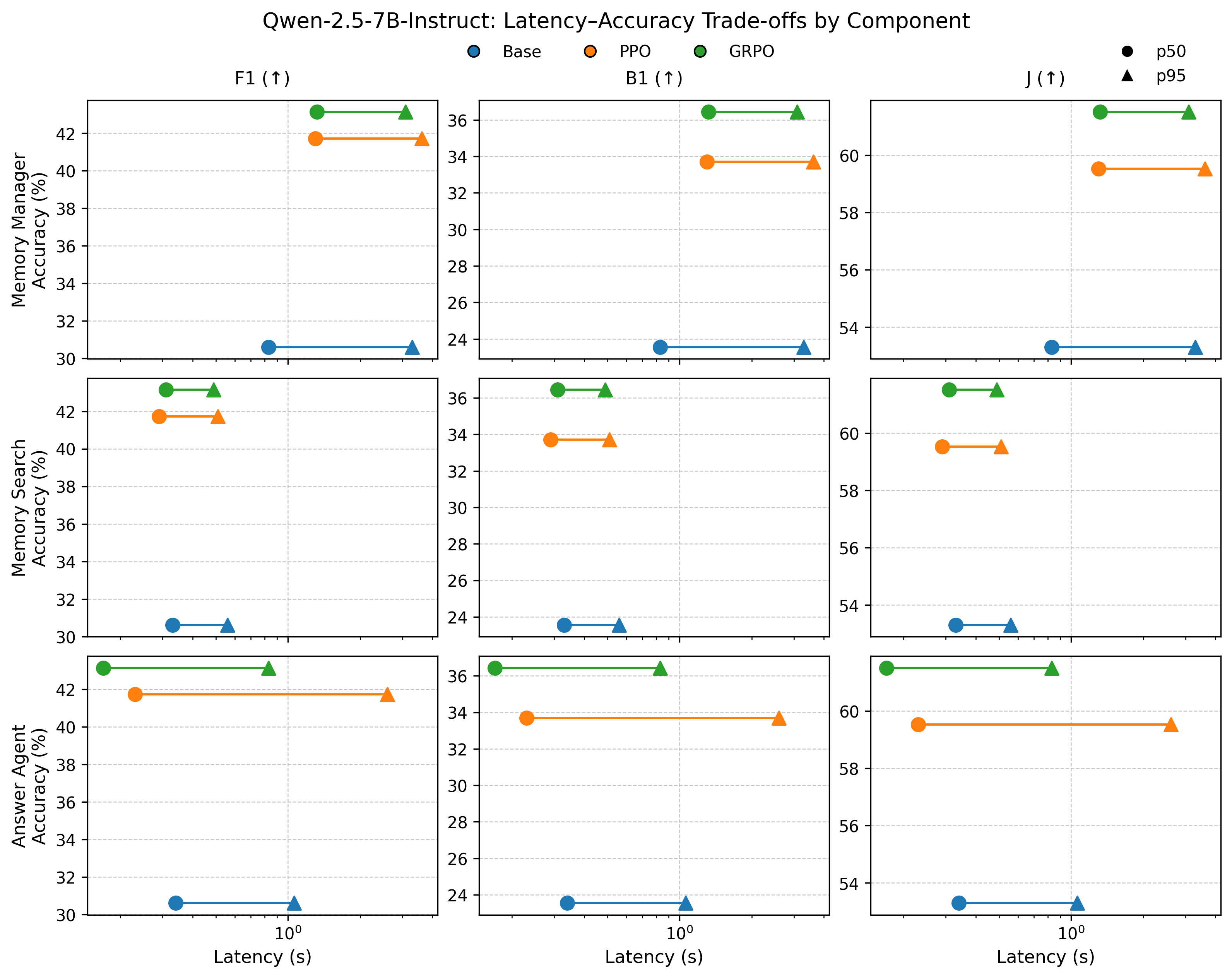} 
\caption{Latency-accuracy comparison across pipeline components on Qwen-2.5-7B-Instruct. Points represent median (p50) and tail (p95) latency versus accuracy for the base model and RL-trained variants (PPO, GRPO).}
\label{fig:latency_accuracy_qwen}
\end{figure}

\paragraph{Accuracy-Latency Relationship}
Figures~\ref{fig:latency_accuracy_llama} and~\ref{fig:latency_accuracy_qwen} further illustrate the relationship between accuracy and latency across components. In contrast to retrieval-heavy pipelines, Memory-R1 achieves higher accuracy while simultaneously reducing both median and tail latency. This behavior indicates a Pareto improvement rather than a trade-off, where learned memory distillation and policy optimization enable the model to reason more efficiently without sacrificing correctness.

Overall, these results demonstrate that Memory-R1 improves inference efficiency in addition to accuracy, especially in the Answer Agent component, and that reinforcement learning can lead to more streamlined reasoning behavior rather than increased computational cost.

\end{document}